\pdfoutput=1

\documentclass[11pt, dvipsnames]{article}
\usepackage{amsmath}
\usepackage[final]{acl}
\usepackage{appendix}

\newcommand{\red}[1]{\textcolor{red}{#1}}

\usepackage{times}
\usepackage{latexsym}
\usepackage{booktabs}
\usepackage{multirow}
\usepackage{multirow}
\usepackage{makecell}
\usepackage{graphicx}
\usepackage{colortbl}
\usepackage[T1]{fontenc}
\usepackage[utf8]{inputenc}
\usepackage{balance}

\usepackage{microtype}
\usepackage{array}
\usepackage{CJKutf8}
\usepackage{inconsolata}
\usepackage{enumitem}
\usepackage{graphicx}
\usepackage{algorithm}
\usepackage{algorithmic}

\title{Can Multiple Responses from an LLM Reveal the Sources of Its Uncertainty?}

\author{
  \textbf{Yang Nan\textsuperscript{1}},
  \textbf{Pengfei He\textsuperscript{2}},
  \textbf{Ravi Tandon\textsuperscript{1}},
  \textbf{Han Xu\textsuperscript{1}}
\\
\\
  \textsuperscript{1}University of Arizona,
  \textsuperscript{2}Michigan State University
\\
\\
  \texttt{\{yangnan, tandonr, xuhan2\}@arizona.edu,}  \texttt{hepengf1@msu.edu}
}

\definecolor{reduction1}{RGB}{198,219,239} 
\definecolor{reduction2}{RGB}{158,202,225} 
\definecolor{reduction3}{RGB}{107,174,214} 

\definecolor{rate1}{RGB}{255,236,217}  
\definecolor{rate2}{RGB}{255,204,153}  
\definecolor{rate3}{RGB}{255,153,102}  

\newcommand{\reductionColor}[1]{%
  \ifdim #1 pt = 0.899 pt \cellcolor{reduction3}#1\relax
  \else\ifdim #1 pt = 0.693 pt \cellcolor{reduction2}#1\relax
  \else\ifdim #1 pt = 0.462 pt \cellcolor{reduction1}#1\relax
  \else\ifdim #1 pt = 0.641 pt \cellcolor{reduction3}#1\relax
  \else\ifdim #1 pt = 0.540 pt \cellcolor{reduction2}#1\relax
  \else\ifdim #1 pt = 0.475 pt \cellcolor{reduction1}#1\relax
  \else\ifdim #1 pt = 0.680 pt \cellcolor{reduction3}#1\relax
  \else\ifdim #1 pt = 0.547 pt \cellcolor{reduction2}#1\relax
  \else\ifdim #1 pt = 0.324 pt \cellcolor{reduction1}#1\relax
  \else\ifdim #1 pt = 0.322 pt \cellcolor{reduction1}#1\relax
  \fi\fi\fi\fi\fi\fi\fi\fi\fi\fi
}

\newcommand{\rateColor}[1]{%
  \ifdim #1 pt = 67.08 pt \cellcolor{rate3}#1\relax
  \else\ifdim #1 pt = 51.69 pt \cellcolor{rate2}#1\relax
  \else\ifdim #1 pt = 53.75 pt \cellcolor{rate3}#1\relax
  \else\ifdim #1 pt = 37.08 pt \cellcolor{rate2}#1\relax
  \else\ifdim #1 pt = 24.30 pt \cellcolor{rate1}#1\relax
  \else\ifdim #1 pt = 28.08 pt \cellcolor{rate1}#1\relax
  \else\ifdim #1 pt = 42.09 pt \cellcolor{rate3}#1\relax
  \else\ifdim #1 pt = 30.20 pt \cellcolor{rate2}#1\relax
  \else\ifdim #1 pt = 32.29 pt \cellcolor{rate2}#1\relax
  \else\ifdim #1 pt = 27.70 pt \cellcolor{rate1}#1\relax
  \else\ifdim #1 pt = 23.72 pt \cellcolor{rate1}#1\relax
  \else\ifdim #1 pt = 23.84 pt \cellcolor{rate1}#1\relax
  \fi\fi\fi\fi\fi\fi\fi\fi\fi\fi\fi\fi
}

\usepackage{array}
\newcolumntype{G}{>{\columncolor{gray!7}}p{1.6cm}}
\newcolumntype{H}{>{\columncolor{gray!15}}p{1.6cm}}

\begin{document}
\maketitle

\begin{abstract}
Large language models (LLMs) have delivered significant breakthroughs across diverse domains but can still produce unreliable or misleading outputs, posing critical challenges for real-world applications. While many recent studies focus on quantifying model uncertainty, relatively little  work has been devoted to \textit{diagnosing the source of uncertainty}.  In this study, we show that, when an LLM is uncertain, the patterns of disagreement among its multiple generated responses contain rich clues about the underlying cause of uncertainty. To illustrate this point, we collect multiple responses from a target LLM and employ an auxiliary LLM to analyze their patterns of disagreement. The auxiliary model is tasked to reason about the likely source of uncertainty, such as whether it stems from ambiguity in the input question, a lack of relevant knowledge, or both. In cases involving knowledge gaps, the auxiliary model also identifies the specific missing facts or concepts contributing to the uncertainty. In our experiment, we validate our framework on AmbigQA, OpenBookQA, and MMLU-Pro, confirming its generality in diagnosing distinct uncertainty sources. Such diagnosis shows the potential for relevant manual interventions that improve LLM performance and reliability.
\end{abstract}

\section{Introduction}

\begin{figure*}[t]
\centering
  \includegraphics[width=0.96\linewidth]{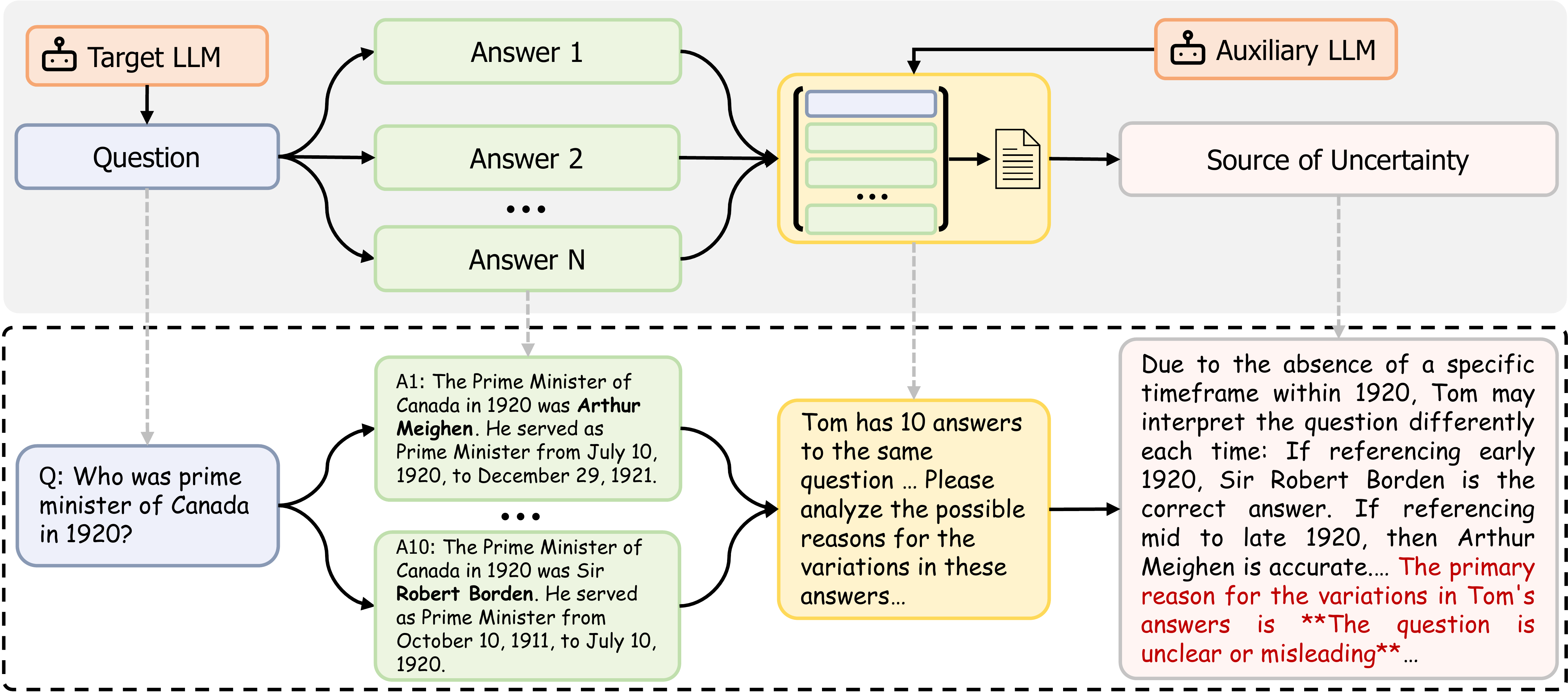} 
  \caption{
  Illustration of the framework and example. We generate multiple responses from a target LLM and use an auxiliary LLM to analyze their disagreement patterns. The top shows the overall process flow, and the bottom presents a concrete example of diagnosing uncertainty for a sample question.}
  \label{fig:figure1}
\end{figure*}



Large Language Models (LLMs) have demonstrated remarkable performance across a wide range of applications, including natural language understanding~\cite{brown2020language,chowdhery2023palm}, reasoning~\cite{wei2022chain,wang2022self}, and decision-making~\cite{chen2021decision,yao2023react}. Ensuring the trustworthiness and reliability of LLMs has become imperative as their capabilities continue to advance. This requirement is particularly critical in sensitive domains such as healthcare~\cite{kung2023performance} and law~\cite{surden2018artificial}, where high uncertainty can lead to significant risks and safety issues~\cite{bommasani2021opportunities}. Accurately quantifying uncertainty helps determine whether a model's prediction can be trusted: low uncertainty indicates a reliable answer, while high uncertainty suggests that the response requires further scrutiny or should be rejected. In the literature, various approaches have been proposed for uncertainty quantification in LLMs, including verbalization-based methods~\cite{kadavath2022language,yin2023large,xiong2023can}, perplexity-based methods~\cite{huang2023look,duan2023shifting}, and self-consistency methods~\cite{wang2022self, yadkori2024believe, xiong2023can}. Among them, self-consistency based methods, which generate multiple independent responses and assess their agreement, have usually demonstrated more promising results, such as a stronger ability to forecast model errors.

Despite these efforts, relatively fewer works have focused on one key challenge: \textit{how to precisely identify the source of uncertainty of LLMs.} In fact, knowing why the model yields highly uncertain responses to a given question is crucial. For example, it can enable the model users or developers to diagnose whether the uncertainty stems from inherent ambiguity in the question or from the model's insufficient knowledge~\cite{hou2023decomposing}. Consequently, this precise diagnosis can later guide targeted improvements: if the uncertainty arises from unclear details of the query, users can refine it; whereas if it results from a lack of specific knowledge of model, developers can upgrade or fine-tune the model with additional data, or users can modify the query to explicitly include the missing knowledge. Notably, similar topics have been explored in traditional models~\cite{kendall2017uncertainties}, but they may not straightforwardly generalize to LLMs, as discussed in Section~\ref{sec:2.2}.

To address the challenge mentioned above, we explore whether multiple responses from LLMs (e.g., obtained during self-consistency assessment) can reveal clues about the source of uncertainty, as they could contain rich contextual information that naturally reflect the underlying cause of uncertainty. As shown in Figure~\ref{fig:figure1}, when asking 
``Who was the prime minister of Canada in 1920?'', some responses indicate the answer is Arthur Meighen but also mention he took office in July. While, others gives the answer as Robert Borden but note that his serving is until July. Analyzing these inconsistencies reveals that each answer understands this question differently by interpreting the term ``in 1920'' differently. It shows that the uncertainty of the model primarily results from unclear question details, rather than a lack of knowledge.


Based on this finding, we explore whether an LLM can automatically diagnose the source of its uncertainty by analyzing patterns of disagreement among its multiple outputs. Refer to the overall framework as illustrated in Figure~\ref{fig:figure1}, we collect the responses from a ``target'' LLM for multiple times, and then employ an ``auxiliary'' LLM to scrutinize these responses and analyze their patterns of disagreement. Specifically, the auxiliary model is prompted to distinguish among various types of uncertainty sources:
(1) whether the uncertainty stems from unclear or under-specified input, or (missing factual or conceptual information), or \textit{Both} (a combination of the two). (2) for samples labeled \textit{Knowledge Gaps} or \textit{Both}, pinpoint the exact factual or conceptual knowledge missing from the reasoning, thereby even more precisely identifying which critical knowledge failure underlies the model's uncertainty.


In our experiment, we first evaluate the approach on the AmbigQA \cite{min2020ambigqa} and OpenBookQA \cite{mihaylov2018can} datasets, both of which contain a variety of fact-based and commonsense questions that are ambiguous or missing key information. We find that representative models 
exhibit notable uncertainty of \textit{Question Ambiguity} on a substantial portion of these. Moreover, if we provide clarification to those questions, we observe that samples labeled as ``Question Ambiguity'' exhibit a great decrease in uncertainty, thereby demonstrating the effectiveness of our uncertainty attribution. In contrast, for questions where the uncertainty stems from missing knowledge, such clarification has little effect and the uncertainty persists. Furthermore, we conduct another study of the Physics and Chemistry subsets of MMLU-Pro ~\cite{wang2024mmlu} which requires various domain knowledge. In these settings, the auxiliary model can successfully identify key missing knowledge components that hinder the target model's performance. Overall, these results suggest the potential to effectively differentiate between distinct sources of uncertainty of LLMs and help guide further appropriate manual interventions.

\section{Related Work}
\subsection{Uncertainty Quantification of LLMs}

Uncertainty plays a critical role in large language models (LLMs). Prior research indicates that LLMs often exhibit overconfidence, raising trust concerns for practical applications~\cite{tian2023just}. Existing uncertainty quantification approaches broadly fall into three categories: 

\paragraph{Verbalization.} 
This class of methods exploits the model's ability to self-report uncertainty by prompting it for confidence judgments (e.g., ``On a scale from 0\% to 100\%, how certain are you?'') and mapping the verbal response to a numerical uncertainty score. \cite{tian2023just,xiong2023can}. Early work demonstrated that GPT-3 could explicitly verbalize its uncertainty~\cite{lin2022teaching}, further studies explored self-awareness across model sizes~\cite{kadavath2022language}, highlighting gaps between model and human uncertainty calibration~\cite{yin2023large}. Recent prompting strategies have further improved uncertainty estimation and model calibration~\cite{tian2023just,xiong2023can}.

\paragraph{Perplexity.} This line of methods quantifies uncertainty using the model's token-level predictive probabilities, where lower perplexity corresponds to higher confidence \cite{huang2023look,duan2023shifting}. Perplexity, initially introduced by~\citet{jelinek1990self}, reflects predictive probability distributions \cite{chen1998evaluation}. \citet{blatz2004confidence} extended perplexity to token-level uncertainty estimation in machine translation, and recent work adopted geometric averaging to mitigate sequence-length biases~\cite{huang2023look,duan2023shifting}.

\paragraph{Self-consistency.} Measure uncertainty by sampling multiple independent Chain-of-Thought responses and quantifying their agreement \cite{wang2022self,yadkori2024believe,xiong2023can,becker2024cycles}. Recent extensions further quantify uncertainty through semantic similarity among responses, such as clustering semantically equivalent sequences~\cite{kuhn2023semantic} or computing covariance between inner states of different responses~\cite{chen2024inside}.  

\subsection{Uncertainty Decomposition}\label{sec:2.2}

Decomposing uncertainty in LLMs is essential for precisely identifying deficiencies at different levels—whether due to inherent data noise or model limitations—thus guiding targeted improvements. Existing studies in the literature typically divide such uncertainty into two categories: \textit{(1) epistemic uncertainty}, which reflects the model's lack of sufficient training data or parameter capacity to generalize correctly, and \textit{(2) aleatoric uncertainty}, which arises from ambiguity in the input \cite{kendall2017uncertainties, hou2023decomposing}. Prior methods such as Bayesian Neural Networks (BNNs)~\cite{neal2012bayesian,hasenclever2017distributed} and Deep Ensembles (DEs)~\cite{lakshminarayanan2017simple} have been used to decompose uncertainty by modeling prediction variability through either posterior sampling or model disagreement. 

However, these approaches are impractical for LLMs because their enormous size makes repeated weight sampling or training multiple model instances prohibitively expensive, and proprietary, black-box APIs prevent access to internal parameter distributions.
Relatively few studies have examined this problem in the context of large language models. One recent work \cite{hou2023decomposing}, with a similar purpose of our study, introduces a method called ``input clarification ensembling'', which first generates multiple clarified variants of a potentially ambiguous prompt and then aggregates the model's outputs over those variants to decompose total uncertainty into its aleatoric and epistemic components. In contrast, our approach infers the source of uncertainty directly from the distribution of generated answers without modifying the original input question. Furthermore, it enables fine-grained attribution by identifying the specific pieces of knowledge that are missing, which is not supported by previous methods.
\section{Preliminary}\label{sec:pre}

In this section, we present a preliminary study comparing the overall accuracy of various uncertainty quantification methods, including {Verbalization} (VERB) \cite{tian2023just,xiong2023can}, {Perplexity} (PPL) \cite{huang2023look,duan2023shifting}, and Self‐Consistency (SC) \cite{wang2022self}. Our findings suggest the \textbf{Self-Consistency} approach generally outperforms alternative methods, positioning it as a promising starting point for investigating the sources of uncertainty.

In detail, we conducted experiments on three benchmarks: GSM8K \cite{cobbe2021training}, MATH \cite{hendrycks2021measuring}, and Natural Questions (NQ) \cite{kwiatkowski2019natural}—using two representative models, Llama3-8B-Instruct \cite{meta2024llama3} and GPT-4o \cite{openai2024hello-gpt4o}. Each method is evaluated using three standard criteria: (1) \textit{Expected Calibration Error (ECE)}: measures the gap between predicted confidence and actual accuracy, indicating how well confidence scores align with correctness \cite{guo2017calibration}. (2) \textit{AUROC}: evaluates the method’s ability to distinguish correct from incorrect answers based on uncertainty scores \cite{fawcett2006introduction}. (3) \textit{Brier Score}: computes the mean squared difference between predicted probabilities and true outcomes, capturing the calibration of uncertainty estimates \cite{brier1950verification} (Further details are provided in the Appendix \ref{sec:appendix1}).
Table~\ref{table:results} presents these results side by side. Across all datasets and both LLMs, Self-Consistency (SC) achieves the lowest ECE, the highest AUROC for mistake detection, and the best (lowest) Brier scores, indicating its strong ability to deliver reliable uncertainty estimates. Accordingly, we adopt Self‐Consistency as our default uncertainty measure for the subsequent analysis. 

\begin{table}[t]
  \centering
  \resizebox{\columnwidth}{!}{
  \begin{tabular}{llcccc}
    \toprule
    Model & Dataset & Method & ECE $\downarrow$ & AUROC $\uparrow$ & Brier $\downarrow$ \\
    \midrule
    \multirow{9}{*}{\makecell[l]{Llama3-8B-\\Instruct}}
      & \multirow{3}{*}{GSM8K}
        & VERB       & 0.146 & 0.636 & 0.182 \\
      &                & PPL        & 0.056 & 0.694 & 0.160 \\
      &                & \textbf{SC} & \textbf{0.054} & \textbf{0.891} & \textbf{0.084} \\
    \cmidrule(l){2-6}
      & \multirow{3}{*}{MATH}
        & VERB       & 0.585 & 0.631 & 0.558 \\
      &                & PPL        & 0.474 & 0.652 & 0.428 \\
      &                & \textbf{SC} & \textbf{0.139} & \textbf{0.723} & \textbf{0.226} \\
    \cmidrule(l){2-6}
      & \multirow{3}{*}{NQ}
        & VERB       & 0.282 & 0.613 & 0.279 \\
      &                & PPL        & 0.548 & 0.520 & 0.523 \\
      &                & \textbf{SC} & \textbf{0.140} & \textbf{0.745} & \textbf{0.187} \\
    \midrule
    \multirow{9}{*}{GPT-4o}
      & \multirow{3}{*}{GSM8K}
        & VERB       & 0.077 & 0.663 & 0.076 \\
      &                & PPL        & –     & –     & –     \\
      &                & \textbf{SC} & \textbf{0.031} & \textbf{0.824} & \textbf{0.047} \\
    \cmidrule(l){2-6}
      & \multirow{3}{*}{MATH}
        & VERB       & 0.585 & 0.631 & 0.558 \\
      &                & PPL        & –     & –     & –     \\
      &                & \textbf{SC} & \textbf{0.182} & \textbf{0.828} & \textbf{0.186} \\
    \cmidrule(l){2-6}
      & \multirow{3}{*}{NQ}
        & VERB       & 0.455     & 0.641     & 0.443   \\
      &                & PPL        & –     & –     & –     \\
      &                & \textbf{SC} & \textbf{0.140}     & \textbf{0.693}     & \textbf{0.118}     \\
    \bottomrule
  \end{tabular}

  }
  \caption{Performance comparison of uncertainty quantification methods. 
  Perplexity-based metrics (PPL) cannot be computed for the black-box GPT-4o model, so it is omitted for GPT-4o.}
  \label{table:results}
\end{table}

\section{Method}

\begin{figure}[!t]
    \includegraphics[width=1\linewidth]{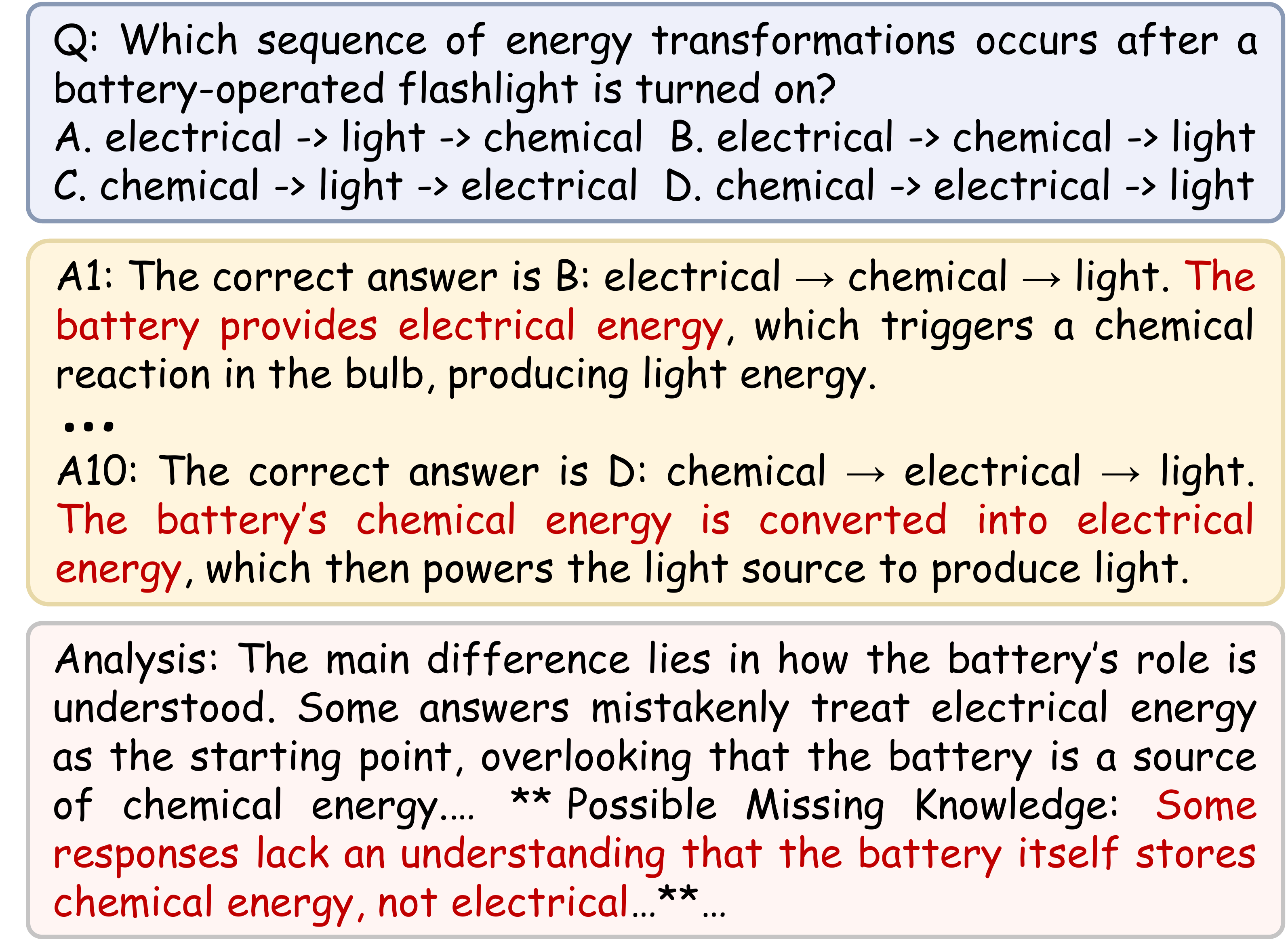}
    \caption{
    An example of using LLM to diagnose uncertainty for a sample question: ten responses were collected, with options B and D each selected five times. To illustrate, two representative responses are shown.
    }
    \label{fig:example}
\end{figure}

Our preliminary evaluation demonstrated that Self‐Consistency produces well‐calibrated uncertainty estimates. Motivated by this finding, we investigate whether the pattern of disagreement among multiple samples can reveal \textbf{\textit{why} a model is uncertain, not merely \textit{how much}}.  


To illustrate, consider the example in Figure~\ref{fig:figure1}, which reflects uncertainty arising from an under-specified question.
For another detailed case, consider the example in Figure~\ref{fig:example}. By aggregating responses for the question ``\textit{Which sequence of energy transformations occurs after a battery-operated flashlight is turned on?}'', we find that the model selects \textit{``electrical -> chemical -> light''} or \textit{``chemical -> electrical -> light}''. Referring to the detailed analysis from the model as shown in Figure~\ref{fig:example}, the primary divergence lies in the model's interpretation of the battery's role. The former one assume that the battery directly contains electrical energy, whereas the latter one correctly recognize that the battery stores chemical energy, which is then converted into electrical energy to power the light. This discrepancy reveals a knowledge gap in the model's understanding of battery function, which underlies its uncertainty. Notably, a manual analysis requires substantial specialized domain knowledge. Therefore, in our work, we leverage this insight and propose a framework that uses an auxiliary LLM to automatically diagnose this source of uncertainty.

\subsection{Notation and Definitions}

To propose our pipeline, we first introduce the necessary notation and definitions. Given a target model $f(\cdot)$ for investigation, we let \(Q\) denote an input question and let \(\{A_1,\dots,A_N\}\) be \(N\) answers sampled from the model. Define the set of unique answers as \(\mathcal{V} = \{v_1, v_2, \dots, v_K\}\). We estimate the probability of each distinct answer \(v_k\) by:
\begin{equation}
P(v_k) \;=\; \frac{1}{N}\sum_{j=1}^{N}\mathbf{1}(A_j = v_k),
\end{equation}
where \(\mathbf{1}(\cdot)\) is the indicator function. Then, the uncertainty of the model to the question \(Q\) can be measured by the Shannon entropy of this distribution~\cite{shannon1948mathematical, wang2022self}:
\begin{equation}
\label{eq:2}
U(Q) \;=\; -\sum_{k=1}^{K} P(v_k)\,\log P(v_k),
\end{equation}
which captures how widely the answers are spread over~$\mathcal{V}$: 
higher entropy indicates greater disagreement among the \(N\) samples and therefore higher uncertainty. Finally, we set a threshold \(\tau\) so any question with \(U(Q) > \tau\) is marked as high‐uncertainty and selected for deeper analysis.

\begin{figure}[!t]
  \includegraphics[width=\columnwidth]{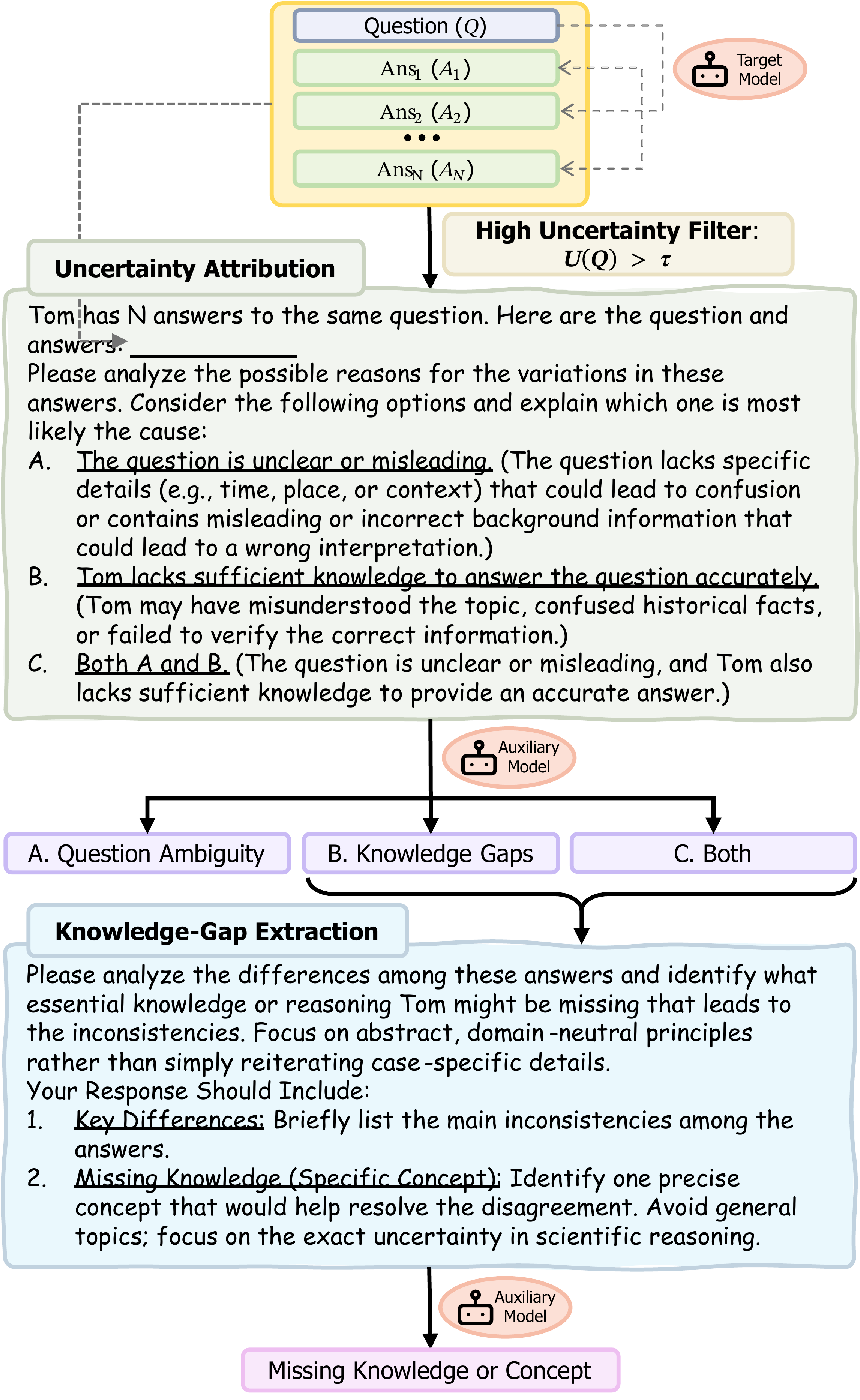}
  \caption{
  Framework of our pipeline for extracting the precise source of uncertainty: (1) Filter high-uncertainty samples; (2) Uncertainty Source Diagnosis: Uncertainty Attribution and Knowledge-Gap Extraction. 
  }
  \label{fig:framework}
\end{figure}

\subsection{Framework}

Given the notation and definitions, we now describe the \textit{two-phase, two-step} pipeline for extracting the precise source of uncertainty\footnotemark. The architecture of our pipeline is illustrated in Figure~\ref{fig:framework}.

\footnotetext{ 
The pseudocode for the entire framework are presented in Appendix \ref{sec:appendix3}. }

\paragraph{Phase I: High-Uncertainty Filtering.} 
For each question \(Q\), we generate \(N\) answers and compute its uncertainty score \(U(Q)\) according to Eq.\ref{eq:2}, which adheres the basic pipeline of Self Consistency for uncertainty estimation. We then select only those samples with \(U(Q)>\tau\) for subsequent analysis, since high‐uncertainty cases indicate possible mistakes (refer Section~\ref{sec:pre}).

\paragraph{Phase II: Two-Step Diagnosis.}
\paragraph{(1)\textit{ Uncertainty Attribution.}} 
Concatenate each filtered question \(Q\) with its \(N\) answers and prompt an auxiliary LLM to analyze and attribute the uncertainty. 
To guide this process, we design a prompt framed around a fictional character (``Tom'') who holds \(N\) answers to the same question (see the prompt in green part of Figure~\ref{fig:framework}). It is because prior work suggests that models reason more reliably when evaluating others' responses rather than their own \cite{lin2022teaching}. This third-person framing helps reduce self-reference bias and encourages more consistent judgments. We extend the conventional two-way decomposition (epistemic and aleatoric uncertainty) into the following:
\begin{enumerate} [itemsep=0.5ex, topsep=1ex]
  \item Question Ambiguity — the question itself is unclear or under-specified, leading to divergent interpretations. This category corresponds to aleatoric uncertainty, as it arises from inputs that allow multiple plausible interpretations due to their inherent vagueness.
  \item Knowledge Gaps — the model fails to retrieve or apply the necessary factual or conceptual information. This replaces the traditional epistemic category.
  \item Both — the case involves both an ambiguous question and a missing knowledge component, jointly causing uncertainty. 
\end{enumerate}



\paragraph{(2) \textit{Knowledge-Gap Extraction.}}
For samples with label \(L\in\{\text{\textit{Knowledge Gaps}}, {\text{\textit{Both}}\}}\), we concatenate the original question \(Q\) with its \(N\) answers and prompt the auxiliary LLM to identify the specific fact or concept that is missing from the response generation, and we denote it as \(K\). See the blue part of Figure~\ref{fig:framework}, the prompt guides the model to analyze the key differences among the responses; and identify the specific piece of missing knowledge that could explain these differences. This module is crucial as it enables us to precisely identify which specific piece of knowledge is missing or misunderstood, in complex reasoning tasks that involve multiple pieces of knowledge (see more discussions in the experiments in Section~\ref{sec:5.2}).
This process is also illustrated in the example shown in Figure~\ref{fig:example}, where the multiple answers of the LLM reveal that the uncertainty stems from an unclear understanding of the ``battery's function'', instead of other concepts. This capability is crucial because it pinpoints the exact missing knowledge. This capability is crucial because it pinpoints the exact missing knowledge, enabling targeted interventions, such as injecting the identified facts into the context to boost model  performance.

\begin{table*}[ht]
\centering
\resizebox{0.96\textwidth}{!}{
\begin{tabular}{lllcccc}
\toprule
\textbf{Dataset} & \textbf{Model} & \textbf{Label} & \textbf{Unc. (Before)} & \textbf{Unc. (After)} & \textbf{Unc. Reduction} & \textbf{Unc. Reduction Rate (\%)} \\
\midrule
\multirow{6}{*}{AmbigQA} & \multirow{3}{*}{\makecell[l]{Llama3-8B\\-Instruct}} & Question Ambiguity & 1.869 & 1.176 & \cellcolor[HTML]{FFC48A}0.693 & \cellcolor[HTML]{8DDDDC}37.08 \\
       &  & Both               & 2.005 & 1.442 & \cellcolor[HTML]{FFDAB0}0.563 & \cellcolor[HTML]{B5EAEA}28.08 \\
       &  & Knowledge Gaps     & 1.902 & 1.440 & \cellcolor[HTML]{FFEBD6}0.462 & \cellcolor[HTML]{DAF5F5}24.30 \\
\cmidrule(lr){2-7}
       & \multirow{3}{*}{\makecell[l]{GPT-3.5\\-turbo}} & Question Ambiguity & 1.522 & 0.881 & \cellcolor[HTML]{FFC48A}0.641 & \cellcolor[HTML]{8DDDDC}42.09 \\
       &  & Both               & 1.673 & 1.133 & \cellcolor[HTML]{FFDAB0}0.540 & \cellcolor[HTML]{B5EAEA}32.29 \\
       &  & Knowledge Gaps     & 1.572 & 1.097 & \cellcolor[HTML]{FFEBD6}0.475 & \cellcolor[HTML]{DAF5F5}30.20 \\
\cmidrule(lr){1-7}
\multirow{6}{*}{OpenbookQA} & \multirow{3}{*}{\makecell[l]{Llama3-8B\\-Instruct}} & Question Ambiguity & 1.340 & 0.441 & \cellcolor[HTML]{FFB877}0.899 & \cellcolor[HTML]{8DDDDC}67.08 \\
           &  & Both               & 1.264 & 0.585 & \cellcolor[HTML]{FFC48A}0.680 & \cellcolor[HTML]{A9E7E6}53.75 \\
           &  & Knowledge Gaps     & 1.058 & 0.511 & \cellcolor[HTML]{FFDAB0}0.547 & \cellcolor[HTML]{B5EAEA}51.69 \\
\cmidrule(lr){2-7}
           & \multirow{3}{*}{\makecell[l]{GPT-3.5\\-turbo}} & Question Ambiguity & 1.171 & 0.846 & \cellcolor[HTML]{FFEBD6}0.324 & \cellcolor[HTML]{DAF5F5}27.70 \\
           &  & Both               & 1.349 & 1.028 & \cellcolor[HTML]{FFF8F0}0.322 & \cellcolor[HTML]{F0FBFB}23.84 \\
           &  & Knowledge Gaps     & 1.214 & 0.926 & \cellcolor[HTML]{FFFDFB}0.288 & \cellcolor[HTML]{F0FBFB}23.72 \\
\bottomrule
\end{tabular}}
\caption{Validation results of \textit{Uncertainty Attribution}. All results are computed on high-uncertainty samples. \textbf{Unc.} denotes model uncertainty. \textbf{Unc. (Before)} and \textbf{Unc. (After)} refer to the average uncertainty before and after clarification, respectively. \textbf{Unc. Reduction} indicates the absolute decrease in uncertainty, while \textbf{Unc. Reduction Rate (\%)} quantifies the relative reduction.}
\label{tab:uncertainty_reduction}
\end{table*}

\begin{table}[t]
\centering
\setlength{\tabcolsep}{2pt}
\resizebox{\columnwidth}{!}{
\begin{tabular}{llcccc}
\toprule
\textbf{Dataset} & \textbf{Label} 
& \makecell[c]{\textbf{Unc.}\\\textbf{(Before)}} 
& \makecell[c]{\textbf{Unc.}\\\textbf{(After)}} 
& \makecell[c]{\textbf{Unc.}\\\textbf{Reduction}} 
& \makecell[c]{\textbf{Unc. Reduct.}\\\textbf{Rate (\%)}} \\
\midrule
\multirow{3}{*}{AmbigQA}
  & Question Ambig. & 1.92 & 1.39 & \cellcolor[HTML]{FFDAB0}0.53 & \cellcolor[HTML]{B5EAEA}27.73 \\
  & Both               & 1.96 & 1.51 & \cellcolor[HTML]{FFEBD6}0.45 & \cellcolor[HTML]{DAF5F5}22.90 \\
  & Knowledge Gaps     & 1.91 & 1.50 & \cellcolor[HTML]{FFEBD6}0.41 & \cellcolor[HTML]{DAF5F5}21.38 \\
\cmidrule(lr){1-6}
\multirow{3}{*}{OpenbookQA}
  & Question Ambig. & 1.37 & 0.44 & \cellcolor[HTML]{FFB877}0.93 & \cellcolor[HTML]{8DDDDC}67.74 \\
  & Both               & 1.13 & 0.41 & \cellcolor[HTML]{FFC48A}0.72 & \cellcolor[HTML]{8DE5E4}63.60 \\
  & Knowledge Gaps     & 1.23 & 0.66 & \cellcolor[HTML]{FFC48A}0.57 & \cellcolor[HTML]{A9E7E6}46.72 \\
\bottomrule
\end{tabular}
}
\caption{Evaluation results using Llama3-8B-Instruct for both generation and analysis. Table format and metrics follow those of Table~\ref{tab:uncertainty_reduction}.}
\label{tab:abalation}
\end{table}

\section{Experiments}
In this section, we present comprehensive experiments to validate the effectiveness of our proposed method for identifying the source of uncertainty in LLMs. Specifically, our experiments focus on answering two core questions:
\begin{enumerate}[label={\textit{(\roman*)}}]
    \item Can the Uncertainty Attribution module accurately distinguish between different sources of uncertainty? (Section \ref{sec:5.1})
    \item Does the Knowledge-Gap Extraction module reliably reveal the knowledge deficiencies in the reasoning process? (Section \ref{sec:5.2})
\end{enumerate}
Unless explicitly stated for certain ablation or replication scenarios, we use GPT o1-mini \cite{openai2024o1mini} as the auxiliary model throughout our experiments, given its strong reasoning capabilities.

\subsection{Validation of Uncertainty Attribution}
\label{sec:5.1}

In this subsection, we evaluate whether the \textit{Uncertainty Attribution} module can effectively distinguish among three sources of uncertainty: \textit{Question Ambiguity}, \textit{Knowledge Gaps}, and \textit{Both}.


\paragraph{Setup.}
We validate the module on benchmarks containing abundant ambiguous questions. AmbigQA~\cite{min2020ambigqa} consists of open-ended Natural Questions that admit multiple valid answers; we evaluate on its 2,002-sample validation set. OpenBookQA~\cite{mihaylov2018can} comprises elementary-level, multiple-choice science questions; we use the first 500 examples from its training set. 
We use Llama3-8B-Instruct and GPT-3.5-turbo \cite{openai2023chatgpt} as target models since they exhibit substantial uncertainty on the two benchmarks, yielding enough high-uncertainty cases for meaningful analysis.
For each question, we generate \(N=10\) answers as a balance between reliable uncertainty estimation and computational cost, following common practice in self-consistency methods, and compute their uncertainty scores. We then apply a threshold \(\tau = 0.89\), chosen to exclude confident cases while retaining enough high-uncertainty samples for analysis.

\paragraph{Experimental Design.} Although datasets like AmbigQA annotate certain questions as ambiguous, these annotations are rather subjective and such questions may not actually be ambiguous to the model. Thus, in our experiment, we instead assess the accuracy of our uncertainty attribution by comparing the reduction in uncertainty before and after clarification across three categories to assess this module's effectiveness. In particular, for each high‐uncertainty question, we generate a clarified version (see Appendix~\ref{sec:appendix4} for the detailed procedure of clarification) \cite{kuhn2022clam,zhang2023clarify}, then sample  \(N\) answers with the target model and recompute uncertainty. If the target LLM's uncertainty drops by a large margin after clarification, it suggests the uncertainty is highly likely due to ``Question Ambiguity''.



\paragraph{Results.}
The experiment results are shown in Table \ref{tab:uncertainty_reduction}.  Clarification leads to the greatest reduction in uncertainty for samples labeled as \textit{Question Ambiguity}, followed by \textit{Both}, and then \textit{Knowledge Gaps}. On AmbigQA, Llama3-8B-Instruct achieves relative uncertainty reductions of 37.1\%, 28.1\%, and 24.3\% for the three categories, while GPT-3.5-turbo yields comparable reductions of 42.1\%, 32.3\%, and 30.2\%. This pattern demonstrates that our attribution aligns with the model's behavior: uncertainty caused by ambiguity is significantly reduced once the question is clarified. Notably, ``Knowledge Gaps'' cases exhibit a modest decrease, since the extra clarification can enhance model understanding to reduce uncertainty. In the only baseline work~\cite{hou2023decomposing}, they also employ the idea of input clarification to attribute uncertainty. Our result in Table \ref{tab:uncertainty_reduction} shows high alignment with their results. Overall, these findings confirm the validity of our \textit{Uncertainty Attribution} module\footnotemark.

\footnotetext{Representative examples of multiple responses and their analyses are provided in Appendix~\ref{sec:appendix5}.}

In our study, the auxiliary model is more advanced than the target model, which may not reflect realistic deployment scenarios. Thus, we replicate the key experiments using a single LLM (Llama3-8B-Instruct) for both answer generation and uncertainty analysis. As shown in Table~\ref{tab:abalation}, the overall trends mirror those obtained with o1-mini: uncertainty reduction still follows the order \textit{Question Ambiguity} > \textit{Both} > \textit{Knowledge Gaps} across both datasets, which reconfirms our conclusion. 

\begin{table*}[t]
\centering
\resizebox{0.96\textwidth}{!}{
\begin{tabular}{lllcccccc}
\toprule
\multirow{2}{*}{\raisebox{-0.5ex}{\textbf{Dataset}}} 
  & \multirow{2}{*}{\raisebox{-0.5ex}{\textbf{Model}}}
  & \multicolumn{2}{c}{\textbf{Before}} 
  & \multicolumn{2}{c}{\textbf{After}}
  & \multirow{2}{*}{\makecell[c]{\raisebox{-0.5ex}{\textbf{Unc. Reduction}}\\\raisebox{-0.5ex}{\textbf{Rate(\%)}}}}
  & \multirow{2}{*}{\makecell[c]{\raisebox{-0.5ex}{\textbf{Acc. Improvement}}\\\raisebox{-0.5ex}{\textbf{Rate(\%)}}}} \\
  \cmidrule(lr){3-4}\cmidrule(lr){5-6}
  & 
  & \textbf{Unc.} & \textbf{Acc. (\%)} 
  & \textbf{Unc.} & \textbf{Acc. (\%)} 
  &  &  \\
\midrule
\multirow{4}{*}{\makecell[l]{MMLU-Pro-Physics}}
  & Llama3-8B-Instruct & 1.83 & 28.29 & 1.59 & 34.78 & \cellcolor{gray!7.5}13.26 & \cellcolor{gray!15}6.49 \\
  & GPT-3.5-turbo      & 1.81 & 39.43 & 1.65 & 43.48 & \cellcolor{gray!7.5}8.97  & \cellcolor{gray!15}4.05 \\
  & GPT-4o             & 1.76 & 29.63 & 0.94 & 72.22 & \cellcolor{gray!7.5}46.51 & \cellcolor{gray!15}42.59 \\
  & o1-mini            & 1.39 & 50.00 & 1.05 & 70.83 & \cellcolor{gray!7.5}24.63 & \cellcolor{gray!15}20.83 \\
\cmidrule(lr){1-8}
\multirow{4}{*}{\makecell[l]{MMLU-Pro-Chemistry}}
  & Llama3-8B-Instruct & 1.90 & 30.77 & 1.64 & 35.90 & \cellcolor{gray!7.5}13.64 & \cellcolor{gray!15}5.13 \\
  & GPT-3.5-turbo      & 1.88 & 41.67 & 1.63 & 50.00 & \cellcolor{gray!7.5}13.45 & \cellcolor{gray!15}8.33 \\
  & GPT-4o             & 1.77 & 37.04 & 0.71 & 74.07 & \cellcolor{gray!7.5}60.01 & \cellcolor{gray!15}37.04 \\
  & o1-mini            & 1.58 & 52.17 & 1.33 & 56.52 & \cellcolor{gray!7.5}15.69 & \cellcolor{gray!15}4.35 \\
\bottomrule
\end{tabular}
}
\caption{Validation results of \textit{Knowledge-Gap Extraction}. All results are computed on high-uncertainty samples. \textbf{Unc.} denotes uncertainty and \textbf{Acc.} denotes accuracy. \textbf{Before} refers to the original performance, while \textbf{After} reflects performance with knowledge added.\textbf{Unc. Reduction Rate} indicates the relative decrease in uncertainty, and \textbf{Acc. Improvement Rate} represents the increase in accuracy.}

\label{tab:knowledge_gap_results}
\end{table*}

\subsection{Validation of Knowledge-Gap Extraction}
\label{sec:5.2}

In this subsection, we check whether \textit{Knowledge-Gap Extraction} module can identify the specific knowledge that contributes to model uncertainty. 

\paragraph{Setup.}
We select two challenging subdomains, Physics and Chemistry, from the MMLU-Pro dataset~\cite{wang2024mmlu}. It is a highly demanding benchmark featuring expert-level, cross-disciplinary multiple-choice questions designed to test advanced reasoning capabilities across professional domains. Each question includes ten answer options, often requiring nuanced understanding and multi-step inference. Moreover, due to the curated nature of MMLU-Pro, the questions are generally well-formed and unambiguous, allowing us to proceed directly with knowledge extraction without performing prior uncertainty attribution. These characteristics make MMLU-Pro a suitable testbed for rigorously evaluating whether our module can identify and compensate for missing knowledge in complex reasoning scenarios. We evaluate our method across four LLMs: Llama3-8B-Instruct, GPT-3.5-turbo, GPT-4o, and o1-mini. All other settings are identical to those in Section~\ref{sec:5.1}.

\paragraph{Experimental Design.}
To validate the effectiveness of our knowledge extraction module, we supplement the original question with relevant ``external knowledge'', which are retrieved based on the missing concept identified by our module. Then, we check whether it can improve the model's performance. Specifically, for each high-uncertainty sample labeled as \textit{Knowledge Gaps} or \textit{Both}, we first extract a concise description of the key missing knowledge. The extracted knowledge phrase is used as a query via the OpenAI web-search tool \cite{openai2025gpt4_1, openai2025websearch} to retrieve a short passage that explains the concept in more detail. For example, if the missing knowledge involves an unclear understanding of how a battery works, we retrieve a brief explanation of battery functionality. The retrieved passage is prepended to the original question as additional context. We then sample \(N\) answers from the target model, compute uncertainty and accuracy after injecting the retrieved knowledge, and compare these metrics to the original results to assess the module's effectiveness. 

\paragraph{Results.}

The validation results are presented in Table~\ref{tab:knowledge_gap_results}. Our method effectively identifies specific knowledge gaps, as evidenced by consistent improvements in both uncertainty and accuracy across all models and datasets after knowledge injection. Particularly notable are the results for GPT-4o, which achieves uncertainty reductions of 46.51\% and 60.01\%, accompanied by accuracy improvements of 42.59\% and 37.04\% on Physics and Chemistry, respectively. Furthermore, when both the target and auxiliary models are instantiated as o1-mini, we still observe significant gains: on the Physics subset, uncertainty decreases by 24.63\% and accuracy increases by 20.83\%, underscoring that our framework's effectiveness derives from its design rather than reliance on any particular model. Additional  results are provided in Appendix~\ref{sec:appendix2}.

\paragraph{Additional Analysis.}
In our result from Table~\ref{tab:knowledge_gap_results}, the additionally retrieved knowledge does not always lead a correct answer. To investigate the reason, we randomly sampled 20 high‐uncertainty questions from MMLU-Pro-Physics and 20 from MMLU-Pro-Chemistry for manual inspection. In the samples, even after knowledge injection, the model still erred on 12 Physics and 11 Chemistry samples. 
Under these errors, we find they usually stem from lapses in logic and the complexity of multi-step calculations 
For an example (which presented in Figure \ref{fig:example4} in Appendix),  when addressing the acid–base pH calculation, our method identifies the Knowledge Gap as ``Stoichiometric Calculations in Acid–Base Reactions''. The incorrect responses exhibit mole–concentration confusion (mole $\leftrightarrow$ concentration; Answers 2, 6), limiting-reagent misidentification (Answers 1, 7), arithmetic/logarithm slip-ups (Answers 3, 10), and pH/pOH formula misuse (Answer 8). 
Similarly, in Figure \ref{fig:example3} the model mis‐uses the phase‐inversion rule and optical‐path‐difference formula, and makes arithmetic or unit‐conversion mistakes.  Under these types of knowledge gaps, the model will still make mistake even supplementary information is provided. In contrast, in cases like those in Figures \ref{fig:example1} and \ref{fig:example2}—where only conceptual knowledge is missing—knowledge injection alone sufficed to correct the model's output. This contrast shows that, although our module reliably diagnoses exactly which fact or principle is missing, solving gaps that require multi‐step quantitative reasoning may demand not just better context but also improvements in the model's inferential and arithmetic capacities. Crucially, these observations do not detract from our method's validity: our primary goal is to identify the source of uncertainty. The detailed error analysis, together with the overall reductions in uncertainty and improvements in accuracy, confirms that our Knowledge‐Gap Extraction module accurately pinpoints the factual deficits driving model uncertainty.
\section{Conclusion}

In this paper, we propose a unified and generalizable framework for diagnosing the source of uncertainty in LLMs, addressing a critical gap in the current literature. By analyzing disagreement across multiple generated answers, our method identifies whether uncertainty arises from question ambiguity, missing specific knowledge during inference, or both. Experiments across diverse models and datasets show that our framework can effectively diagnose the source of uncertainty. In particular, its ability to pinpoint missing knowledge elements offers a new perspective for improving reasoning performance in LLMs. Overall, precisely diagnosing these uncertainty sources enables targeted interventions that reduce uncertainty, bolster model trustworthiness, and facilitate reliable deployment in sensitive, high‐stakes applications.

\newpage
\section{Limitations}

\paragraph{Inference Cost.} One limitation of our framework lies in its inference cost. Since each question requires sampling multiple responses (\(N=10\)) and then running two rounds of auxiliary analysis (Uncertainty Attribution and Knowledge-Gap Extraction), the total number of model invocations can be substantial. This repeated generation and classification may limit the method’s scalability in latency-sensitive or resource-constrained environments, such as real-time applications or deployment on edge devices.

\paragraph{Lack of Direct Evaluation Metrics.}  
Because diagnosing the source of uncertainty in LLMs is a relatively new task, there are no established quantitative metrics for (i) the accuracy of the uncertainty labels produced in the first step, nor for (ii) the precision of the extracted knowledge in the second step. We considered using a separate LLM to score or validate these outputs, but LLM-based evaluation is itself subjective, highly sensitive to prompt design, and often unreliable for fine-grained judgments. Manual annotation could help, but it introduces human subjectivity and does not scale. Instead, we validated the first step by measuring differential uncertainty reduction across the three label categories after input clarification. We validated the second step by measuring performance gains after injecting retrieved knowledge. The first validation cannot provide a precise measure of attribution accuracy because it relies on indirect behavioral signals rather than ground-truth labels. The second validation may understate the true value of the extracted information because models sometimes fail to fully comprehend the provided context, a limitation driven both by their reasoning capacity and by the inclusion of relatively long passages that dilute focus on the key facts. Nonetheless, in the absence of established benchmarks for uncertainty diagnosis, our combined behavioral and performance-based evaluation remains the most rigorous and objective framework currently available.  

\section{Acknowledgment}

Yang Nan and Han Xu is supported by Department of Electrical and Computer Engineering at University of Arizona. Ravi Tandon is supported by NSF grants CCF-2100013, CNS-2209951, CNS-1822071, CNS-2317192, and by the U.S. Department of Energy, Office of Science, Office of Advanced Scientific Computing under Award Number DE-SC-ERKJ422, and NIH Award R01-CA261457-01A1.

\bibliography{reference}

\begin{thebibliography}{46}
\providecommand{\natexlab}[1]{#1}

\bibitem[{Becker and Soatto(2024)}]{becker2024cycles}
Evan Becker and Stefano Soatto. 2024.
\newblock Cycles of thought: Measuring llm confidence through stable explanations.
\newblock \emph{arXiv preprint arXiv:2406.03441}.

\bibitem[{Blatz et~al.(2004)Blatz, Fitzgerald, Foster, Gandrabur, Goutte, Kulesza, Sanchis, and Ueffing}]{blatz2004confidence}
John Blatz, Erin Fitzgerald, George Foster, Simona Gandrabur, Cyril Goutte, Alex Kulesza, Alberto Sanchis, and Nicola Ueffing. 2004.
\newblock Confidence estimation for machine translation.
\newblock In \emph{Coling 2004: Proceedings of the 20th international conference on computational linguistics}, pages 315--321.

\bibitem[{Bommasani et~al.(2021)Bommasani, Hudson, Adeli, Altman, Arora, von Arx, Bernstein, Bohg, Bosselut, Brunskill et~al.}]{bommasani2021opportunities}
Rishi Bommasani, Drew~A Hudson, Ehsan Adeli, Russ Altman, Simran Arora, Sydney von Arx, Michael~S Bernstein, Jeannette Bohg, Antoine Bosselut, Emma Brunskill, and 1 others. 2021.
\newblock On the opportunities and risks of foundation models.
\newblock \emph{arXiv preprint arXiv:2108.07258}.

\bibitem[{Brier(1950)}]{brier1950verification}
Glenn~W Brier. 1950.
\newblock Verification of forecasts expressed in terms of probability.
\newblock \emph{Monthly weather review}, 78(1):1--3.

\bibitem[{Brown et~al.(2020)Brown, Mann, Ryder, Subbiah, Kaplan, Dhariwal, Neelakantan, Shyam, Sastry, Askell et~al.}]{brown2020language}
Tom Brown, Benjamin Mann, Nick Ryder, Melanie Subbiah, Jared~D Kaplan, Prafulla Dhariwal, Arvind Neelakantan, Pranav Shyam, Girish Sastry, Amanda Askell, and 1 others. 2020.
\newblock Language models are few-shot learners.
\newblock \emph{Advances in neural information processing systems}, 33:1877--1901.

\bibitem[{Chen et~al.(2024)Chen, Liu, Chen, Gu, Wu, Tao, Fu, and Ye}]{chen2024inside}
Chao Chen, Kai Liu, Ze~Chen, Yi~Gu, Yue Wu, Mingyuan Tao, Zhihang Fu, and Jieping Ye. 2024.
\newblock Inside: Llms' internal states retain the power of hallucination detection.
\newblock \emph{arXiv preprint arXiv:2402.03744}.

\bibitem[{Chen et~al.(2021)Chen, Lu, Rajeswaran, Lee, Grover, Laskin, Abbeel, Srinivas, and Mordatch}]{chen2021decision}
Lili Chen, Kevin Lu, Aravind Rajeswaran, Kimin Lee, Aditya Grover, Misha Laskin, Pieter Abbeel, Aravind Srinivas, and Igor Mordatch. 2021.
\newblock Decision transformer: Reinforcement learning via sequence modeling.
\newblock \emph{Advances in neural information processing systems}, 34:15084--15097.

\bibitem[{Chen et~al.(1998)Chen, Beeferman, and Rosenfeld}]{chen1998evaluation}
Stanley~F Chen, Douglas Beeferman, and Roni Rosenfeld. 1998.
\newblock Evaluation metrics for language models.

\bibitem[{Chowdhery et~al.(2023)Chowdhery, Narang, Devlin, Bosma, Mishra, Roberts, Barham, Chung, Sutton, Gehrmann et~al.}]{chowdhery2023palm}
Aakanksha Chowdhery, Sharan Narang, Jacob Devlin, Maarten Bosma, Gaurav Mishra, Adam Roberts, Paul Barham, Hyung~Won Chung, Charles Sutton, Sebastian Gehrmann, and 1 others. 2023.
\newblock Palm: Scaling language modeling with pathways.
\newblock \emph{Journal of Machine Learning Research}, 24(240):1--113.

\bibitem[{Cobbe et~al.(2021)Cobbe, Kosaraju, Bavarian, Chen, Jun, Kaiser, Plappert, Tworek, Hilton, Nakano et~al.}]{cobbe2021training}
Karl Cobbe, Vineet Kosaraju, Mohammad Bavarian, Mark Chen, Heewoo Jun, Lukasz Kaiser, Matthias Plappert, Jerry Tworek, Jacob Hilton, Reiichiro Nakano, and 1 others. 2021.
\newblock Training verifiers to solve math word problems.
\newblock \emph{arXiv preprint arXiv:2110.14168}.

\bibitem[{Duan et~al.(2023)Duan, Cheng, Wang, Zavalny, Wang, Xu, Kailkhura, and Xu}]{duan2023shifting}
Jinhao Duan, Hao Cheng, Shiqi Wang, Alex Zavalny, Chenan Wang, Renjing Xu, Bhavya Kailkhura, and Kaidi Xu. 2023.
\newblock Shifting attention to relevance: Towards the predictive uncertainty quantification of free-form large language models.
\newblock \emph{arXiv preprint arXiv:2307.01379}.

\bibitem[{Fawcett(2006)}]{fawcett2006introduction}
Tom Fawcett. 2006.
\newblock An introduction to roc analysis.
\newblock \emph{Pattern recognition letters}, 27(8):861--874.

\bibitem[{Guo et~al.(2017)Guo, Pleiss, Sun, and Weinberger}]{guo2017calibration}
Chuan Guo, Geoff Pleiss, Yu~Sun, and Kilian~Q Weinberger. 2017.
\newblock On calibration of modern neural networks.
\newblock In \emph{International conference on machine learning}, pages 1321--1330. PMLR.

\bibitem[{Hasenclever et~al.(2017)Hasenclever, Webb, Lienart, Vollmer, Lakshminarayanan, Blundell, and Teh}]{hasenclever2017distributed}
Leonard Hasenclever, Stefan Webb, Thibaut Lienart, Sebastian Vollmer, Balaji Lakshminarayanan, Charles Blundell, and Yee~Whye Teh. 2017.
\newblock Distributed bayesian learning with stochastic natural gradient expectation propagation and the posterior server.
\newblock \emph{Journal of Machine Learning Research}, 18(106):1--37.

\bibitem[{Hendrycks et~al.(2021)Hendrycks, Burns, Kadavath, Arora, Basart, Tang, Song, and Steinhardt}]{hendrycks2021measuring}
Dan Hendrycks, Collin Burns, Saurav Kadavath, Akul Arora, Steven Basart, Eric Tang, Dawn Song, and Jacob Steinhardt. 2021.
\newblock Measuring mathematical problem solving with the math dataset.
\newblock \emph{arXiv preprint arXiv:2103.03874}.

\bibitem[{Hou et~al.(2023)Hou, Liu, Qian, Andreas, Chang, and Zhang}]{hou2023decomposing}
Bairu Hou, Yujian Liu, Kaizhi Qian, Jacob Andreas, Shiyu Chang, and Yang Zhang. 2023.
\newblock Decomposing uncertainty for large language models through input clarification ensembling.
\newblock \emph{arXiv preprint arXiv:2311.08718}.

\bibitem[{Huang et~al.(2023)Huang, Song, Wang, Zhao, Chen, Juefei-Xu, and Ma}]{huang2023look}
Yuheng Huang, Jiayang Song, Zhijie Wang, Shengming Zhao, Huaming Chen, Felix Juefei-Xu, and Lei Ma. 2023.
\newblock Look before you leap: An exploratory study of uncertainty measurement for large language models.
\newblock \emph{arXiv preprint arXiv:2307.10236}.

\bibitem[{Jelinek(1990)}]{jelinek1990self}
Fred Jelinek. 1990.
\newblock Self-organized language modeling for speech recognition.
\newblock \emph{Readings in speech recognition}, pages 450--506.

\bibitem[{Kadavath et~al.(2022)Kadavath, Conerly, Askell, Henighan, Drain, Perez, Schiefer, Hatfield-Dodds, DasSarma, Tran-Johnson et~al.}]{kadavath2022language}
Saurav Kadavath, Tom Conerly, Amanda Askell, Tom Henighan, Dawn Drain, Ethan Perez, Nicholas Schiefer, Zac Hatfield-Dodds, Nova DasSarma, Eli Tran-Johnson, and 1 others. 2022.
\newblock Language models (mostly) know what they know.
\newblock \emph{arXiv preprint arXiv:2207.05221}.

\bibitem[{Kendall and Gal(2017)}]{kendall2017uncertainties}
Alex Kendall and Yarin Gal. 2017.
\newblock What uncertainties do we need in bayesian deep learning for computer vision?
\newblock \emph{Advances in neural information processing systems}, 30.

\bibitem[{Kuhn et~al.(2022)Kuhn, Gal, and Farquhar}]{kuhn2022clam}
Lorenz Kuhn, Yarin Gal, and Sebastian Farquhar. 2022.
\newblock Clam: Selective clarification for ambiguous questions with generative language models.
\newblock \emph{arXiv preprint arXiv:2212.07769}.

\bibitem[{Kuhn et~al.(2023)Kuhn, Gal, and Farquhar}]{kuhn2023semantic}
Lorenz Kuhn, Yarin Gal, and Sebastian Farquhar. 2023.
\newblock Semantic uncertainty: Linguistic invariances for uncertainty estimation in natural language generation.
\newblock \emph{arXiv preprint arXiv:2302.09664}.

\bibitem[{Kung et~al.(2023)Kung, Cheatham, Medenilla, Sillos, De~Leon, Elepa{\~n}o, Madriaga, Aggabao, Diaz-Candido, Maningo et~al.}]{kung2023performance}
Tiffany~H Kung, Morgan Cheatham, Arielle Medenilla, Czarina Sillos, Lorie De~Leon, Camille Elepa{\~n}o, Maria Madriaga, Rimel Aggabao, Giezel Diaz-Candido, James Maningo, and 1 others. 2023.
\newblock Performance of chatgpt on usmle: potential for ai-assisted medical education using large language models.
\newblock \emph{PLoS digital health}, 2(2):e0000198.

\bibitem[{Kwiatkowski et~al.(2019)Kwiatkowski, Palomaki, Redfield, Collins, Parikh, Alberti, Epstein, Polosukhin, Devlin, Lee et~al.}]{kwiatkowski2019natural}
Tom Kwiatkowski, Jennimaria Palomaki, Olivia Redfield, Michael Collins, Ankur Parikh, Chris Alberti, Danielle Epstein, Illia Polosukhin, Jacob Devlin, Kenton Lee, and 1 others. 2019.
\newblock Natural questions: a benchmark for question answering research.
\newblock \emph{Transactions of the Association for Computational Linguistics}, 7:453--466.

\bibitem[{Lakshminarayanan et~al.(2017)Lakshminarayanan, Pritzel, and Blundell}]{lakshminarayanan2017simple}
Balaji Lakshminarayanan, Alexander Pritzel, and Charles Blundell. 2017.
\newblock Simple and scalable predictive uncertainty estimation using deep ensembles.
\newblock \emph{Advances in neural information processing systems}, 30.

\bibitem[{Lin et~al.(2022)Lin, Hilton, and Evans}]{lin2022teaching}
Stephanie Lin, Jacob Hilton, and Owain Evans. 2022.
\newblock \href {https://openreview.net/forum?id=8s8K2UZGTZ} {Teaching models to express their uncertainty in words}.
\newblock \emph{Transactions on Machine Learning Research}.

\bibitem[{{Meta AI}(2024)}]{meta2024llama3}
{Meta AI}. 2024.
\newblock \href {https://github.com/facebookresearch/llama} {Llama-3 documentation}.

\bibitem[{Mihaylov et~al.(2018)Mihaylov, Clark, Khot, and Sabharwal}]{mihaylov2018can}
Todor Mihaylov, Peter Clark, Tushar Khot, and Ashish Sabharwal. 2018.
\newblock Can a suit of armor conduct electricity? a new dataset for open book question answering.
\newblock \emph{arXiv preprint arXiv:1809.02789}.

\bibitem[{Min et~al.(2020)Min, Michael, Hajishirzi, and Zettlemoyer}]{min2020ambigqa}
Sewon Min, Julian Michael, Hannaneh Hajishirzi, and Luke Zettlemoyer. 2020.
\newblock Ambigqa: Answering ambiguous open-domain questions.
\newblock \emph{arXiv preprint arXiv:2004.10645}.

\bibitem[{Neal(2012)}]{neal2012bayesian}
Radford~M Neal. 2012.
\newblock \emph{Bayesian learning for neural networks}, volume 118.
\newblock Springer Science \& Business Media.

\bibitem[{{OpenAI}(2023)}]{openai2023chatgpt}
{OpenAI}. 2023.
\newblock \href {https://openai.com/blog/chatgpt} {Chatgpt (gpt-3.5-turbo) release notes}.

\bibitem[{{OpenAI}(2024{\natexlab{a}})}]{openai2024hello-gpt4o}
{OpenAI}. 2024{\natexlab{a}}.
\newblock {Hello GPT-4o}.
\newblock \url{https://openai.com/index/hello-gpt-4o/}.
\newblock Accessed: 2025-05-20.

\bibitem[{{OpenAI}(2024{\natexlab{b}})}]{openai2024o1mini}
{OpenAI}. 2024{\natexlab{b}}.
\newblock \href {https://platform.openai.com/docs/models/o1-mini} {o1‑mini model card}.

\bibitem[{{OpenAI}(2025{\natexlab{a}})}]{openai2025gpt4_1}
{OpenAI}. 2025{\natexlab{a}}.
\newblock {Introducing GPT-4.1 in the API}.
\newblock \url{https://openai.com/blog/introducing-gpt-4-1-in-the-api}.
\newblock Accessed: 2025-05-17.

\bibitem[{{OpenAI}(2025{\natexlab{b}})}]{openai2025websearch}
{OpenAI}. 2025{\natexlab{b}}.
\newblock {Web Search Documentation}.
\newblock \url{https://platform.openai.com/docs/guides/web-search}.
\newblock Accessed: 2025-05-17.

\bibitem[{Shannon(1948)}]{shannon1948mathematical}
Claude~E Shannon. 1948.
\newblock A mathematical theory of communication.
\newblock \emph{The Bell system technical journal}, 27(3):379--423.

\bibitem[{Surden(2018)}]{surden2018artificial}
Harry Surden. 2018.
\newblock Artificial intelligence and law: An overview.
\newblock \emph{Ga. St. UL Rev.}, 35:1305.

\bibitem[{Tian et~al.(2023)Tian, Mitchell, Zhou, Sharma, Rafailov, Yao, Finn, and Manning}]{tian2023just}
Katherine Tian, Eric Mitchell, Allan Zhou, Archit Sharma, Rafael Rafailov, Huaxiu Yao, Chelsea Finn, and Christopher~D Manning. 2023.
\newblock Just ask for calibration: Strategies for eliciting calibrated confidence scores from language models fine-tuned with human feedback.
\newblock \emph{arXiv preprint arXiv:2305.14975}.

\bibitem[{Wang et~al.(2022)Wang, Wei, Schuurmans, Le, Chi, Narang, Chowdhery, and Zhou}]{wang2022self}
Xuezhi Wang, Jason Wei, Dale Schuurmans, Quoc Le, Ed~Chi, Sharan Narang, Aakanksha Chowdhery, and Denny Zhou. 2022.
\newblock Self-consistency improves chain of thought reasoning in language models.
\newblock \emph{arXiv preprint arXiv:2203.11171}.

\bibitem[{Wang et~al.(2024)Wang, Ma, Zhang, Ni, Chandra, Guo, Ren, Arulraj, He, Jiang et~al.}]{wang2024mmlu}
Yubo Wang, Xueguang Ma, Ge~Zhang, Yuansheng Ni, Abhranil Chandra, Shiguang Guo, Weiming Ren, Aaran Arulraj, Xuan He, Ziyan Jiang, and 1 others. 2024.
\newblock Mmlu-pro: A more robust and challenging multi-task language understanding benchmark.
\newblock In \emph{The Thirty-eight Conference on Neural Information Processing Systems Datasets and Benchmarks Track}.

\bibitem[{Wei et~al.(2022)Wei, Wang, Schuurmans, Bosma, Xia, Chi, Le, Zhou et~al.}]{wei2022chain}
Jason Wei, Xuezhi Wang, Dale Schuurmans, Maarten Bosma, Fei Xia, Ed~Chi, Quoc~V Le, Denny Zhou, and 1 others. 2022.
\newblock Chain-of-thought prompting elicits reasoning in large language models.
\newblock \emph{Advances in neural information processing systems}, 35:24824--24837.

\bibitem[{Xiong et~al.(2023)Xiong, Hu, Lu, Li, Fu, He, and Hooi}]{xiong2023can}
Miao Xiong, Zhiyuan Hu, Xinyang Lu, Yifei Li, Jie Fu, Junxian He, and Bryan Hooi. 2023.
\newblock Can llms express their uncertainty? an empirical evaluation of confidence elicitation in llms.
\newblock \emph{arXiv preprint arXiv:2306.13063}.

\bibitem[{Yadkori et~al.(2024)Yadkori, Kuzborskij, Gy{\"o}rgy, and Szepesv{\'a}ri}]{yadkori2024believe}
Yasin~Abbasi Yadkori, Ilja Kuzborskij, Andr{\'a}s Gy{\"o}rgy, and Csaba Szepesv{\'a}ri. 2024.
\newblock To believe or not to believe your llm.
\newblock \emph{arXiv preprint arXiv:2406.02543}.

\bibitem[{Yao et~al.(2023)Yao, Zhao, Yu, Du, Shafran, Narasimhan, and Cao}]{yao2023react}
Shunyu Yao, Jeffrey Zhao, Dian Yu, Nan Du, Izhak Shafran, Karthik Narasimhan, and Yuan Cao. 2023.
\newblock React: Synergizing reasoning and acting in language models.
\newblock In \emph{International Conference on Learning Representations (ICLR)}.

\bibitem[{Yin et~al.(2023)Yin, Sun, Guo, Wu, Qiu, and Huang}]{yin2023large}
Zhangyue Yin, Qiushi Sun, Qipeng Guo, Jiawen Wu, Xipeng Qiu, and Xuanjing Huang. 2023.
\newblock Do large language models know what they don't know?
\newblock \emph{arXiv preprint arXiv:2305.18153}.

\bibitem[{Zhang and Choi(2023)}]{zhang2023clarify}
Michael~JQ Zhang and Eunsol Choi. 2023.
\newblock Clarify when necessary: Resolving ambiguity through interaction with lms.
\newblock \emph{arXiv preprint arXiv:2311.09469}.

\end{thebibliography}

\appendix

\section{Preliminary Evaluation Details}
\label{sec:appendix1}

\subsection{Uncertainty Quantification Methods}

\paragraph{Verbalization}  
Given input question $x$ and a single model response $\hat y = \mathcal M(x)$, we prompt:
\begin{quote}
\texttt{Question: ``\,$x$''}\\
\texttt{Answer: ``\,$\hat y$''}\\
\texttt{Provide the reasoning correctness probability for the answer.}
\end{quote}
The model’s numeric reply $u\in[0,1]$ is taken as the verbalization confidence score $p^{\mathrm{VERB}}_\theta(\hat y\mid x)$.

\paragraph{Perplexity}  
Let the answer $\hat y=[t_1,\dots,t_m]$ be the model’s token sequence (excluding any end‐of‐sequence token). We collect the likelihood of each token under its conditional context,
\[
\ell_i = p_\theta(t_i\mid x, t_{<i}),
\]
and define the geometric‐mean confidence
\[
p^{\mathrm{PPL}}_\theta(\hat y\mid x)
= \exp\Bigl(\frac{1}{m}\sum_{i=1}^m \ln \ell_i\Bigr).
\]

\paragraph{Self‐Consistency}  
For each input \(x\), we sample \(n\) independent answers \(\{\hat y_i\}_{i=1}^n\sim\mathcal M(\cdot\mid x)\). Let
\[
\hat y^* \;=\;\arg\max_{a}\bigl|\{\,i:\,\hat y_i = a\}\bigr|
\]
be the most frequent answer, and let \(f^* = |\{\,i:\,\hat y_i = \hat y^*\}|\) denote its count. We then define the self‐consistency confidence as
\[
p^{\mathrm{SC}}_\theta(x)
=\frac{f^*}{n},
\]
i.e.\ the relative frequency of the majority answer among the \(n\) samples.  

\subsection{Experimental Setup}

We evaluate on three benchmarks by selecting the first 300 examples of GSM8K and MATH, and the first 200 examples of Natural Questions. For Verbalization and Perplexity, we generate one response per question. For Self-Consistency, we draw $N=10$ samples per question 
to estimate $p^{\mathrm{SC}}$. All experiments use Llama3-8B-Instruct and GPT-4o.

\subsection{Evaluation Metrics}

\paragraph{Expected Calibration Error (ECE)}  
Partition predictions into $K$ confidence bins $\{B_k\}$ and compute
\[
\mathrm{ECE}
= \sum_{k=1}^K \frac{|B_k|}{N}\bigl|\mathrm{acc}(B_k)-\mathrm{conf}(B_k)\bigr|,
\]
where $\mathrm{acc}(B_k)$ is the empirical accuracy and $\mathrm{conf}(B_k)$ the average confidence in bin $k$~\cite{guo2017calibration}.

\paragraph{AUROC}  
Compute the Area Under the Receiver Operating Characteristic curve by ranking predictions by uncertainty and measuring true/false positive rates~\cite{fawcett2006introduction}.

\paragraph{Brier Score}  
For each example $i$, let $u_i$ be the predicted confidence and $y_i\in\{0,1\}$ the correctness indicator. Then
\[
\mathrm{BS}
= \frac{1}{N}\sum_{i=1}^N (u_i - y_i)^2,
\]
which captures both calibration and sharpness of the uncertainty estimates~\cite{brier1950verification}.
\section{Additional Results}
\label{sec:appendix2}

\begin{table*}[t]
\centering
\resizebox{\textwidth}{!}{
\begin{tabular}{lllcccccc}
\toprule
\multirow{2}{*}{\raisebox{-0.5ex}{\textbf{Dataset}}} 
  & \multirow{2}{*}{\raisebox{-0.5ex}{\textbf{Model}}}
  & \multicolumn{2}{c}{\textbf{Before}} 
  & \multicolumn{2}{c}{\textbf{After}}
 & \multirow{2}{*}{\makecell[c]{\raisebox{-0.5ex}{\textbf{Unc. Reduction}}\\\raisebox{-0.5ex}{\textbf{Rate(\%)}}}}
& \multirow{2}{*}{\makecell[c]{\raisebox{-0.5ex}{\textbf{Acc. Improvement}}\\\raisebox{-0.5ex}{\textbf{Rate(\%)}}}} \\
  \cmidrule(lr){3-4}\cmidrule(lr){5-6}
  & 
  & \textbf{Unc.} & \textbf{Acc. (\%)} 
  & \textbf{Unc.} & \textbf{Acc. (\%)} 
  &  &  \\
\midrule
\multirow{3}{*}{\makecell[l]{MMLU-Pro-Physics}}
  & Llama3-8B-Instruct & 1.83 & 28.29 & 1.58 & 35.60 & \cellcolor{gray!7.5}13.90 & \cellcolor{gray!15}7.31 \\
  & GPT-3.5-turbo      & 1.87 & 37.61 & 1.57 & 43.28 & \cellcolor{gray!7.5}15.92 & \cellcolor{gray!15}5.67 \\
  & GPT-4o             & 1.75 & 35.72 & 1.15 & 50.00 & \cellcolor{gray!7.5}34.35 & \cellcolor{gray!15}14.28 \\
\cmidrule(lr){1-8}
\multirow{3}{*}{\makecell[l]{MMLU-Pro-Chemistry}}
  & Llama3-8B-Instruct & 1.83 & 30.71 & 1.61 & 34.64 & \cellcolor{gray!7.5}12.04 & \cellcolor{gray!15}3.93 \\
  & GPT-3.5-turbo      & 1.85 & 39.68 & 1.68 & 40.71 & \cellcolor{gray!7.5}9.52  & \cellcolor{gray!15}1.03 \\
  & GPT-4o             & 1.76 & 34.94 & 1.66 & 35.31 & \cellcolor{gray!7.5}5.29  & \cellcolor{gray!15}0.38 \\
\cmidrule(lr){1-8}
\multirow{3}{*}{\makecell[l]{MMLU-Pro-Law}}
  & Llama3-8B-Instruct & 1.35 & 18.09 & 0.67 & 23.21 & \cellcolor{gray!7.5}50.01 & \cellcolor{gray!15}5.12 \\
  & GPT-3.5-turbo      & 1.42 & 26.73 & 0.88 & 30.50 & \cellcolor{gray!7.5}38.11 & \cellcolor{gray!15}3.76 \\
  & GPT-4o             & 1.41 & 40.63 & 0.82 & 44.72 & \cellcolor{gray!7.5}58.27 & \cellcolor{gray!15}4.09 \\
\bottomrule
\end{tabular}
}
\caption{Validation results of \textit{Knowledge-Gap Extraction}. All results are computed on high-uncertainty samples. \textbf{Unc.} denotes uncertainty and \textbf{Acc.} denotes accuracy. \textbf{Before} refers to the original performance, while \textbf{After} reflects performance with knowledge added.\textbf{Unc. Reduction Rate} indicates the relative decrease in uncertainty, and \textbf{Acc. Improvement Rate} represents the increase in accuracy.}

\label{tab:knowledge_gap_results2}
\end{table*}

\subsection{Experimental Setup}  

In the main experiments (Section \ref{sec:5.2}), retrieved knowledge were obtained via web search. To evaluate whether prompt-based context synthesis can serve as a viable alternative to external retrieval, we instead generate the missing knowledge context directly with o1-mini using a concise prompt (Figure \ref{fig:prompt3}). We conduct these ablations on the Physics, Chemistry, and Law subsets of MMLU-Pro. All other settings remain the same: sample \(N=10\) answers per question; apply the same uncertainty threshold \(\tau=0.89\); and prepend the generated context before re-evaluating uncertainty and accuracy.

\subsection{Results}  
Table~\ref{tab:knowledge_gap_results2} presents the prompt-only validation of our Knowledge-Gap Extraction module. Across the three MMLU-Pro sub-domains, we again observe clear improvements in both uncertainty and accuracy after context injection. In the Physics subset, uncertainty falls by 13.90\%–34.35\% and accuracy rises by 7.31\%–14.28\%; in Chemistry, uncertainty decreases by 5.29\%–12.04\% with accuracy gains of 0.38\%–3.93\%; and in Law, uncertainty is reduced by 38.11\%–58.27\% while accuracy improves by 3.76\%–5.12\%. These results confirm that—even when missing knowledge is synthesized via prompt rather than retrieved externally—our module remains effective at diagnosing and mitigating knowledge deficiencies to reduce uncertainty and boost model performance.

\section{Pseudocode of Framework}
\label{sec:appendix3}
The complete procedure for our two-phase, two-step uncertainty diagnosis framework is detailed in Algorithm~\ref{alg:uncertainty}. Given a test set of questions \( D = \{Q^{(i)}\}_{i=1}^M \), the model first generates \( N \) independent answers per question using stochastic decoding. Each question is then assigned an uncertainty score \( U(Q^{(i)}) \) computed as described in Eq.~\ref{eq:2}. Only samples with \( U(Q^{(i)}) > \tau \) are retained for further analysis, as low-uncertainty cases offer limited diagnostic value.

For each retained question, we apply a structured diagnostic process consisting of two steps. In the first step—\textit{Uncertainty Attribution}—we prompt the LLM to identify whether the cause of uncertainty arises from ambiguity in the question, a knowledge gap, or both. This classification is produced using a third-person prompt format to reduce self-reference bias. In the second step—\textit{Knowledge-Gap Extraction}—we prompt the LLM to extract the specific missing fact or concept \( K^{(i)} \) that would resolve the observed inconsistency. This step is applied only when the uncertainty is attributed to a knowledge gap or both causes. The final output of the pipeline includes an uncertainty label \( L^{(i)} \) for each high-uncertainty question and, when applicable, a corresponding knowledge snippet \( K^{(i)} \).
\label{sec:appendix3}
\begin{algorithm}[!htb]
    \renewcommand{\algorithmicrequire}{\textbf{Input:}}
    \renewcommand{\algorithmicensure}{\textbf{Output:}}
    \caption{Pipeline for Uncertainty Diagnosis}
    \label{alg:uncertainty}
    \begin{algorithmic}[1]
        \REQUIRE Test set $D=\{Q^{(i)}\}_{i=1}^M$, threshold $\tau$, number of samples $N$
        \ENSURE For each $Q^{(i)}$: uncertainty label $L^{(i)}$ and, if applicable, knowledge snippet $K^{(i)}$
        \FOR{each $Q^{(i)}\in D$}
            \STATE $\{A^{(i)}_j\}_{j=1}^N 
            \xleftarrow{\mathrm{LLM}} 
            Q^{(i)}$
            \STATE Compute uncertainty $U(Q^{(i)})$ via Eq.~\ref{eq:2}
            \IF{$U(Q^{(i)})>\tau$}
                \STATE 
                $L^{(i)} 
                \xleftarrow{\mathrm{LLM}} 
                \text{Prompt}_{\mathrm{UA}}\bigl(
                  Q^{(i)}, \{A^{(i)}_j\}
                \bigr)$
                \\ \textcolor{MidnightBlue}{// Uncertainty Attribution}

                \IF{$L^{(i)} \neq \text{\textit{Question Ambiguity}}$}
                    \STATE
                    $K^{(i)} 
                    \xleftarrow{\mathrm{LLM}} 
                    \text{Prompt}_{\mathrm{KGE}}\bigl(
                      Q^{(i)}, \{A^{(i)}_j\}
                    \bigr)$
                    \hfill \textcolor{MidnightBlue}{// Knowledge-Gap Extraction}
                \ENDIF
            \ENDIF
        \ENDFOR
    \end{algorithmic}
\end{algorithm}

\section{Prompts}
\label{sec:appendix4}

\begin{figure*}[t]
    \centering
  \includegraphics[width=0.9\linewidth]{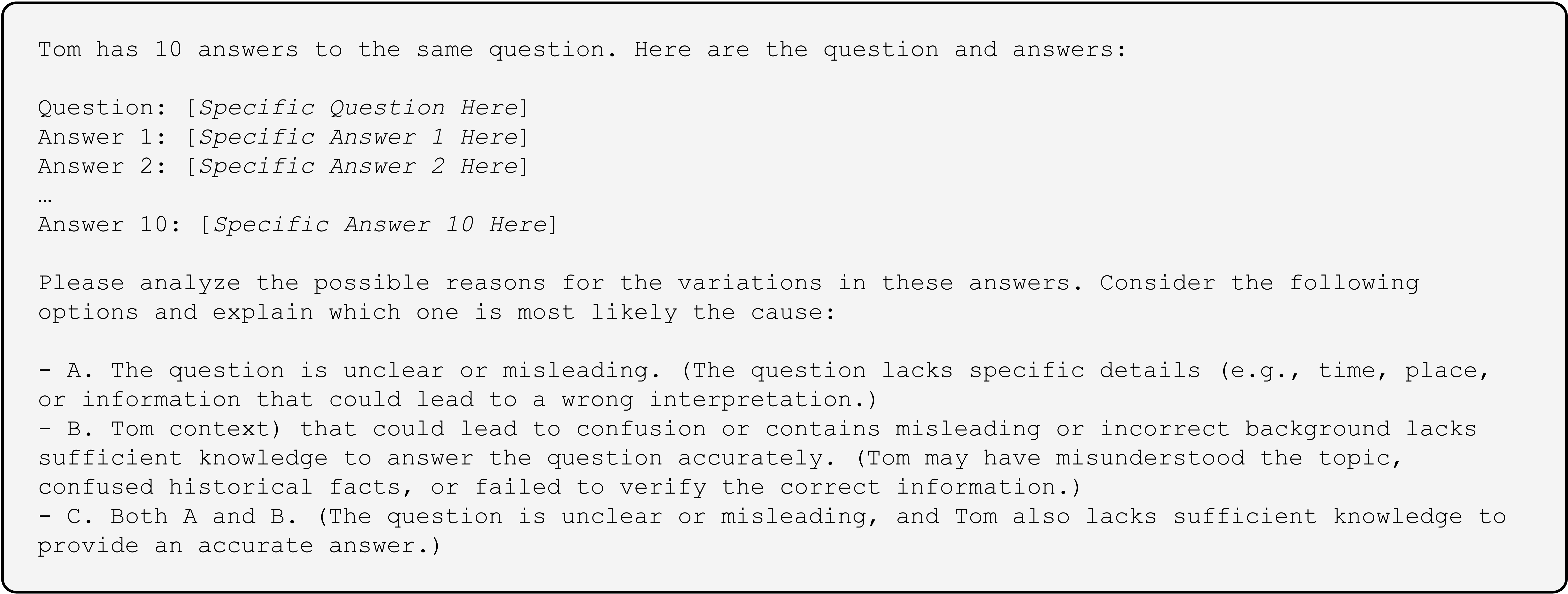} 
  \caption{Prompt template for the \textit{Uncertainty Attribution} step via multi-answer analysis.
  }
  \label{fig:prompt1}
\end{figure*}
\begin{figure*}[t]
\centering
  \includegraphics[width=0.9\linewidth]{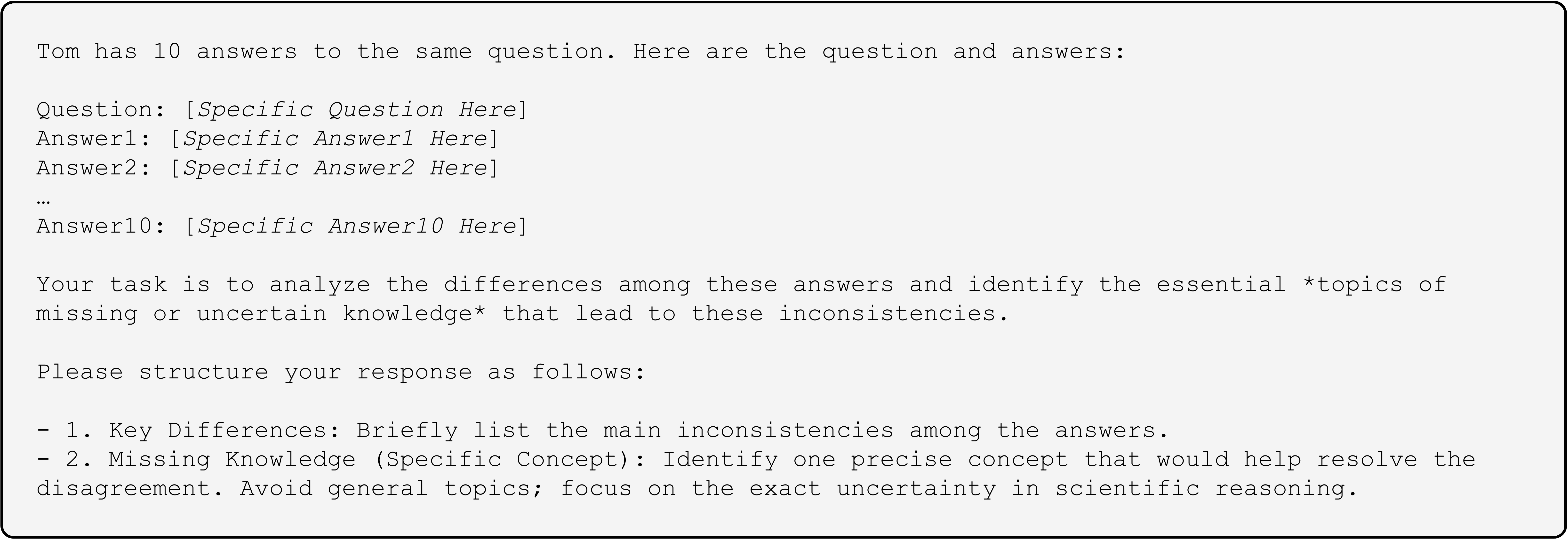} 
  \caption{Prompt template for the \textit{Knowledge-Gap Extraction} step via multi-answer analysis.
  }
  \label{fig:prompt2}
\end{figure*}
\begin{figure*}[t]
 \centering 
  \includegraphics[width=0.9\linewidth]{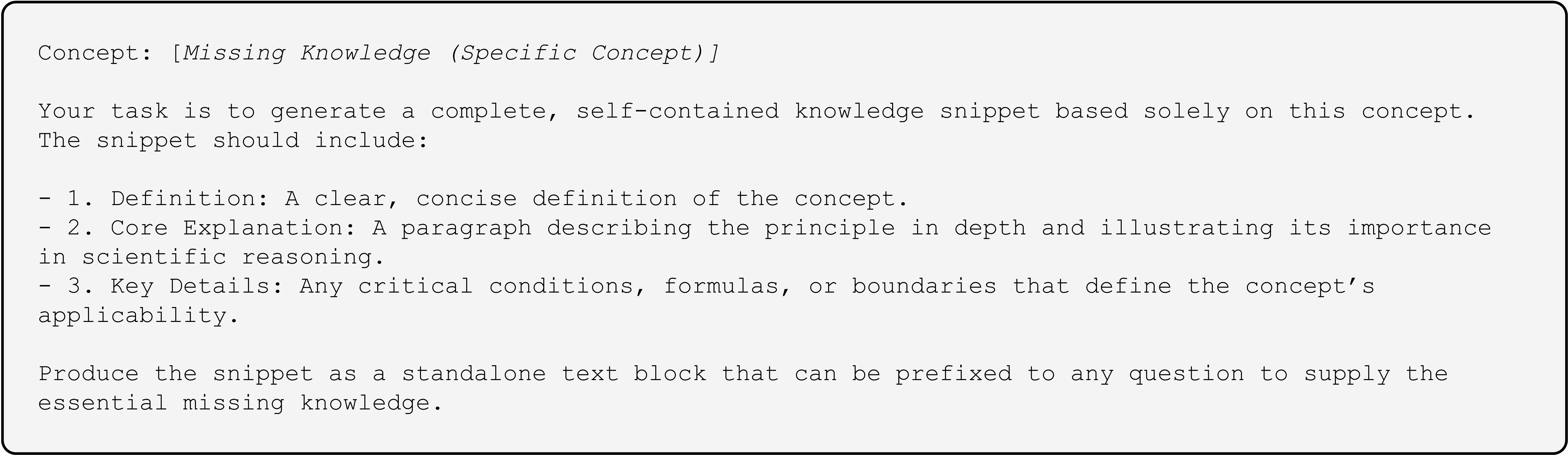} 
  \caption{Prompt template for synthesizing a standalone ``knowledge snippet'' for a given concept.
  }
  \label{fig:prompt3}
\end{figure*}
\begin{figure*}[t]
 \centering 
  \includegraphics[width=0.9\linewidth]{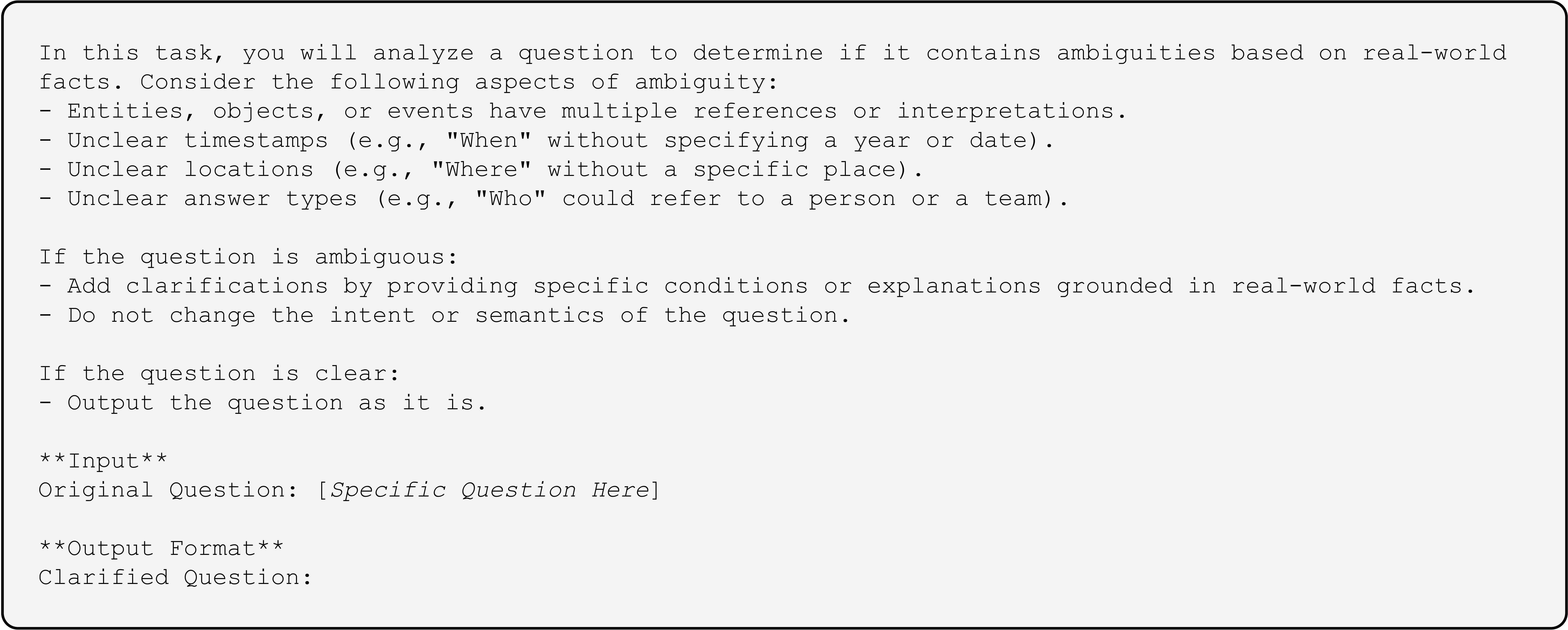} 
  \caption{Prompt template for the Input Clarification step, used to detect and resolve ambiguities.}
  \label{fig:prompt4}
\end{figure*}

We provide the full prompt templates used for each stage of our framework below. Each prompt is carefully designed to guide the model through a structured diagnostic or generation process; complete examples and formatting details are as follows:
\begin{enumerate}
\item Uncertainty Attribution Prompt (Figure~\ref{fig:prompt1}): Frames the task around a fictional character (``Tom'') who offers multiple answers, asks the auxiliary LLM to compare these responses, and choose among ``Question Ambiguity,'' ``Knowledge Gaps,'' or ``Both'' as the source of disagreement. This third-person setup reduces self-reference bias and encourages consistent classification.
\item Knowledge-Gap Extraction Prompt (Figure~\ref{fig:prompt2}): Instructs the auxiliary LLM to first summarize key differences across the sampled answers and then pinpoint the single, precise piece of missing factual or conceptual knowledge that would resolve the inconsistency. The prompt explicitly breaks the task into two steps—difference analysis and knowledge identification—to ensure clarity and focus. 
\item Knowledge Synthesis Prompt (Figure~\ref{fig:prompt3}): Takes a concise knowledge keyword or concept identified in the previous step and instructs o1-mini to generate a self-contained explanatory snippet. This snippet includes a clear definition, core explanation, and any critical conditions or formulas, formatted as a standalone block that can be prefixed to any question as supplemental context. It is only employed in the appendix \ref{sec:appendix2} experiments.
\item Input Clarification Prompt (Figure~\ref{fig:prompt4}): Guides the model to detect real-world ambiguities in the original question—such as unspecified timeframes, locations, or referents—and, if needed, to inject succinct, fact-grounded clarifications without altering the question’s intent. If the question is already clear, the prompt simply reproduces it unchanged.
\end{enumerate}

\section{Examples}
\label{sec:appendix5}

\begin{figure*}[t]
\centering
  \includegraphics[width=0.98\linewidth]{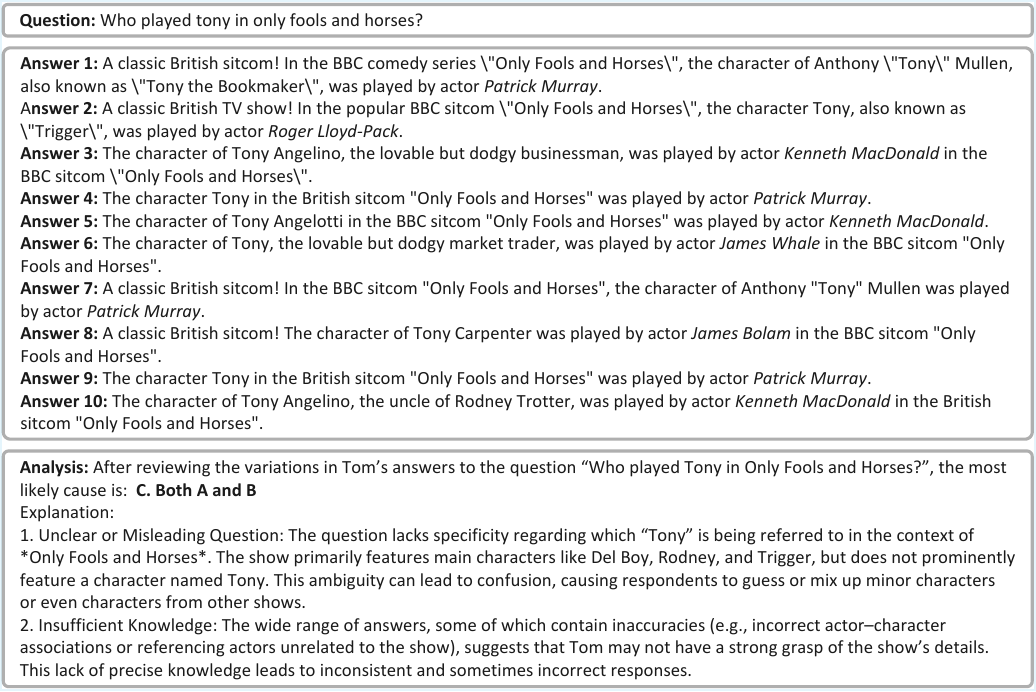} 
  \caption{Example from an AmbigQA sample where the \textit{Uncertainty Attribution module} pinpoints the source of uncertainty.}
  \label{fig:example5}
\end{figure*}
\begin{figure*}[t]
\centering
  \includegraphics[width=0.98\linewidth]{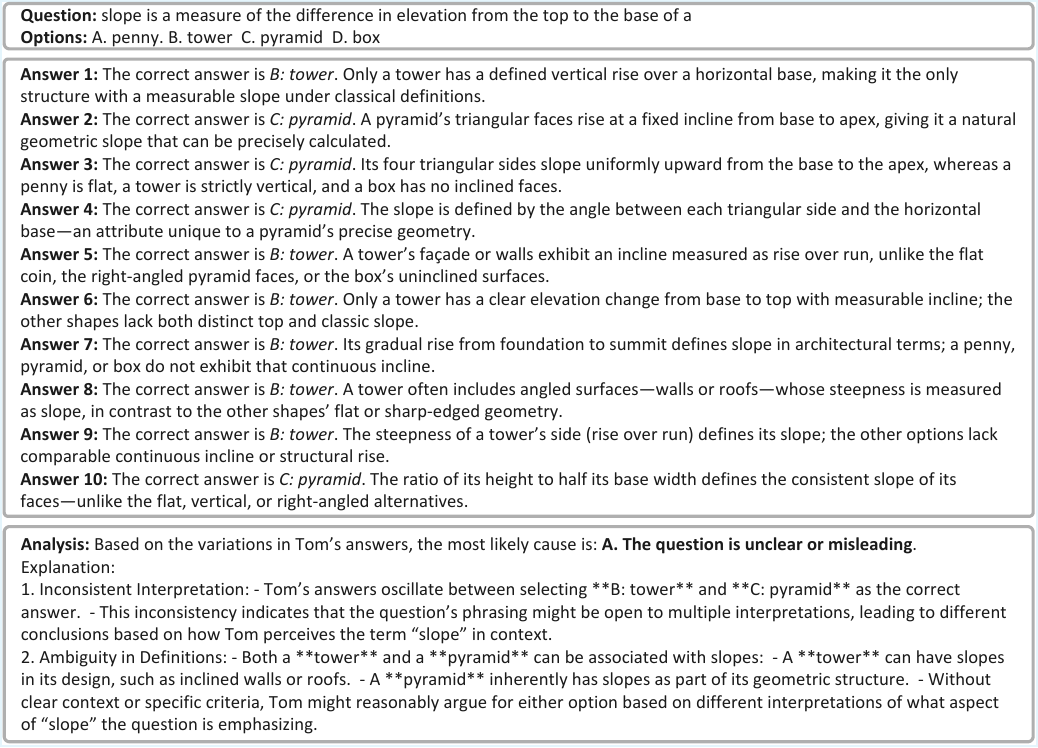} 
  \caption{Example from an OpenBookQA sample where the \textit{Uncertainty Attribution module} pinpoints the source of uncertainty.}
  \label{fig:example6}
\end{figure*}
\begin{figure*}[t]
\centering
  \includegraphics[width=0.98\linewidth]{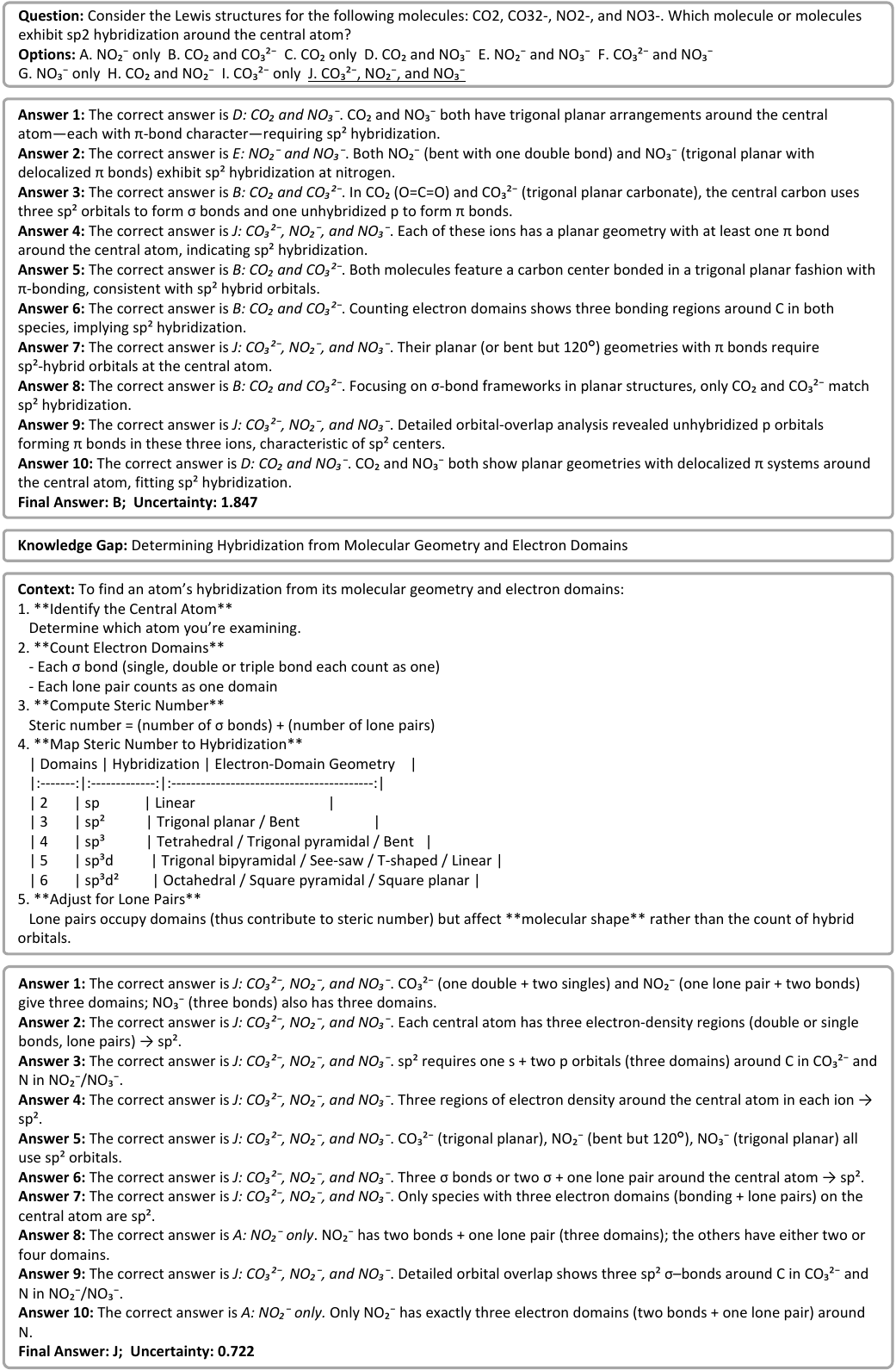} 
  \caption{Example from an MMLU-Pro-Chemistry sample where the model makes a \red{\textbf{correct}} prediction after incorporating external knowledge.}
  \label{fig:example2}
\end{figure*}
\begin{figure*}[t]
\centering
  \includegraphics[width=0.98\linewidth]{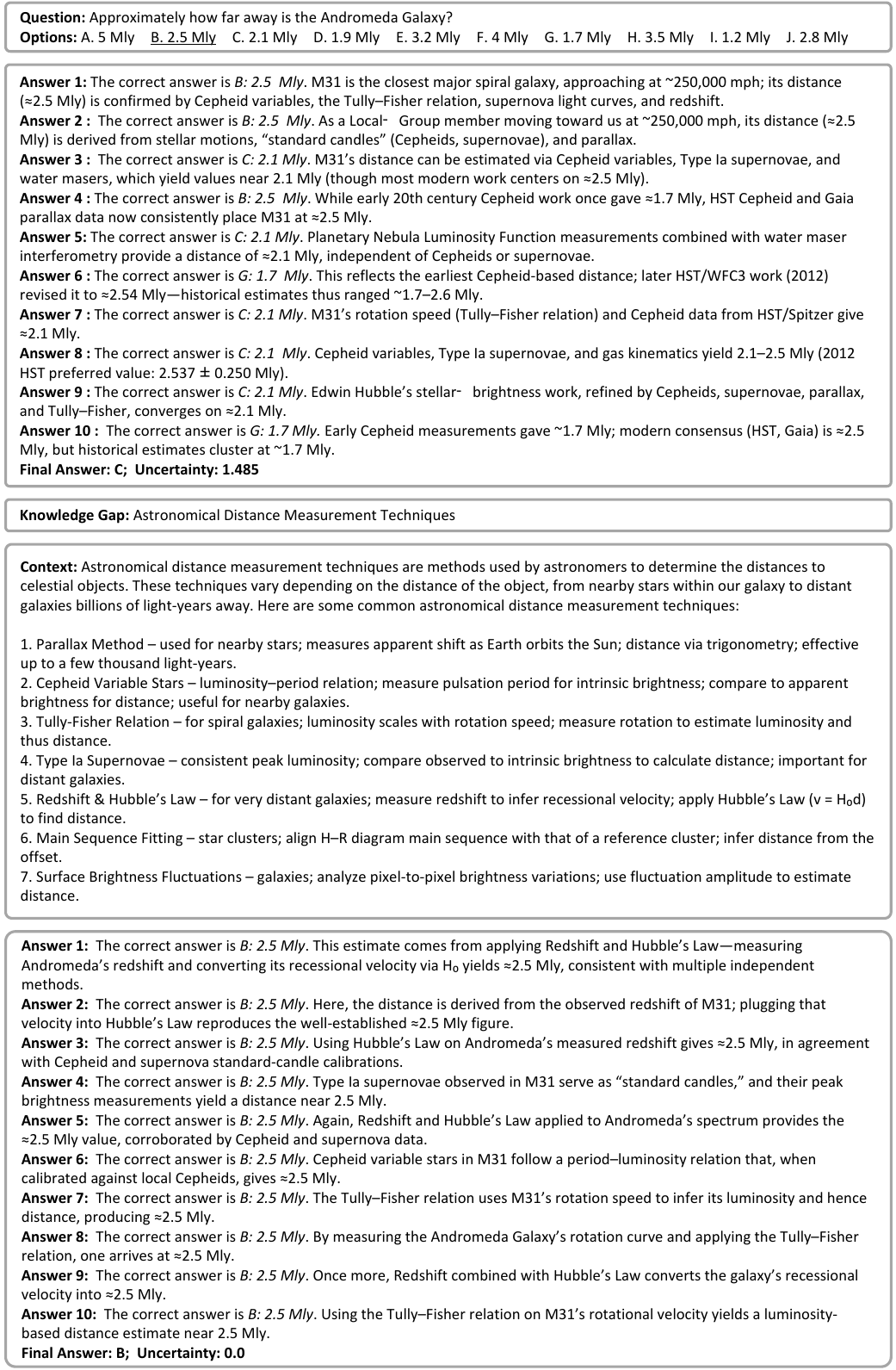} 
  \caption{Example from an MMLU-Pro-Physics sample where the model makes a \red{\textbf{correct}} prediction after incorporating external knowledge.}
  \label{fig:example1}
\end{figure*}
\begin{figure*}[t]
\centering
  \includegraphics[width=0.98\linewidth]{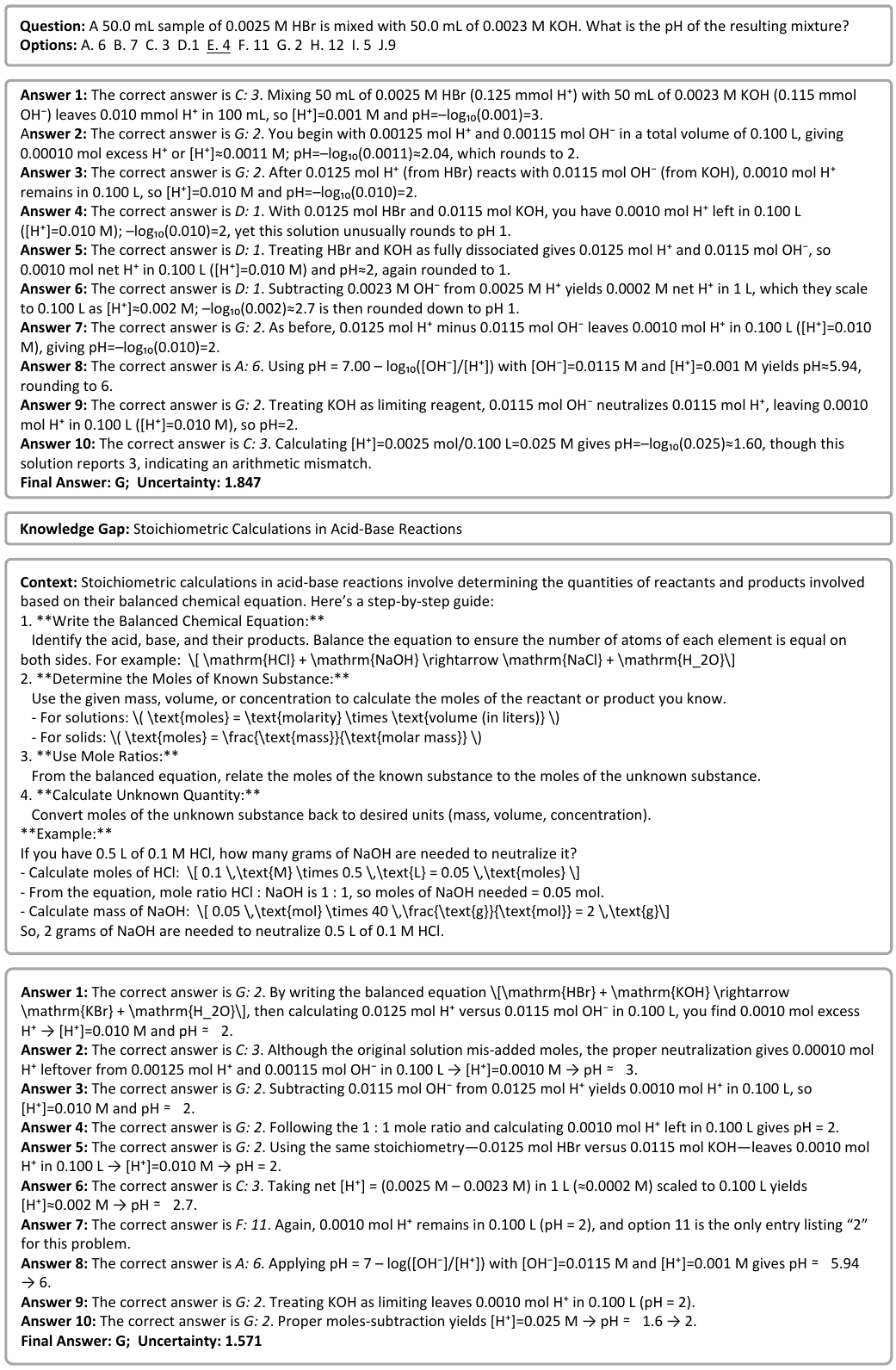} 
  \caption{Example from an MMLU-Pro-Chemistry sample where the model makes a \red{\textbf{incorrect}} prediction after incorporating external knowledge}
  \label{fig:example4}
\end{figure*}
\begin{figure*}[t]
\centering
  \includegraphics[width=0.98\linewidth]{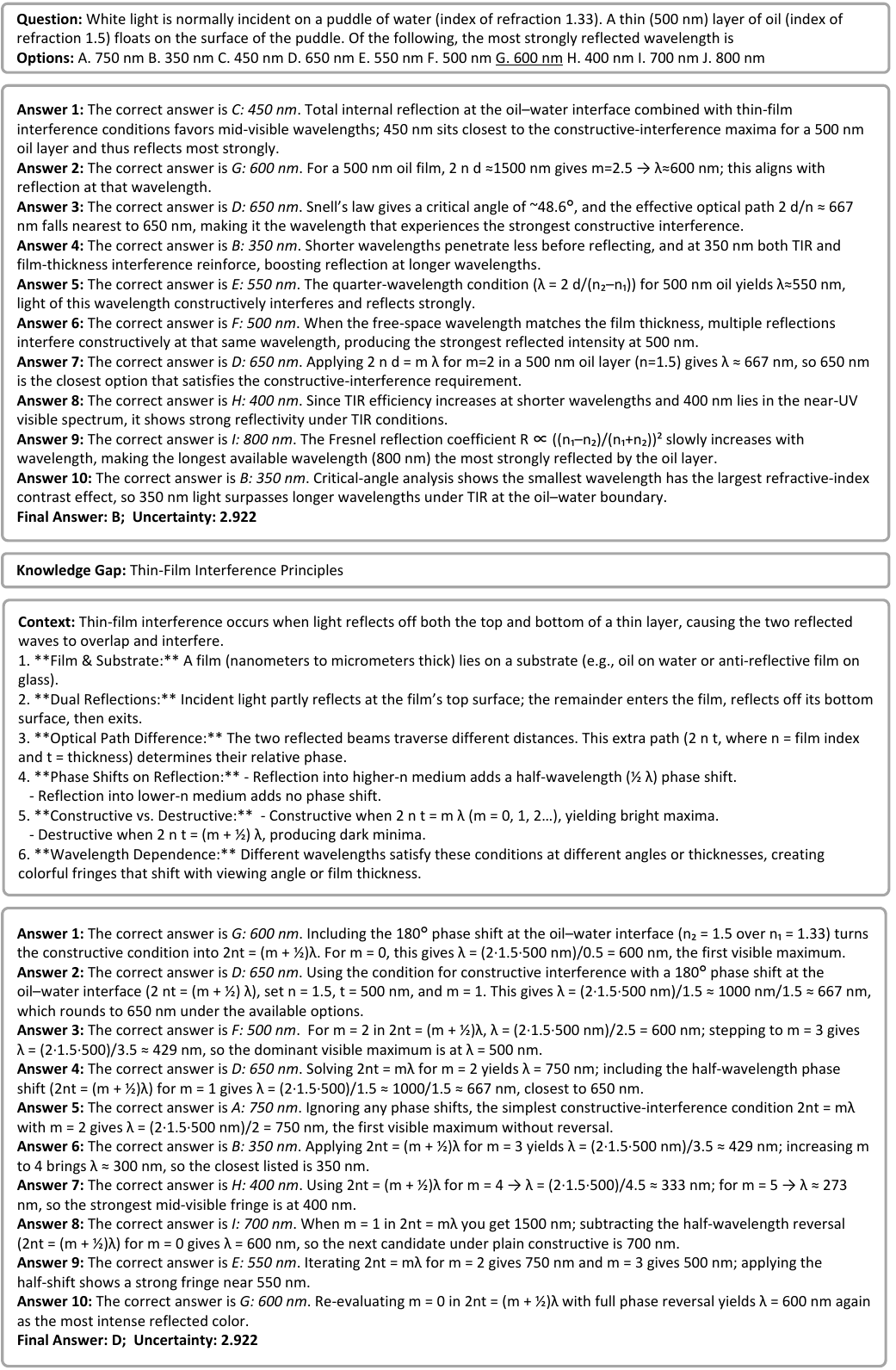} 
  \caption{Example from an MMLU-Pro-Physics sample where the model makes a \red{\textbf{incorrect}} prediction after incorporating external knowledge}
  \label{fig:example3}
\end{figure*}

We present six examples (text truncated for brevity). Figures \ref{fig:example5} and \ref{fig:example6} illustrate the Uncertainty Attribution step. By comparing multiple responses, our method correctly classifies one AmbigQA question as \textit{Both} (under‐specified phrasing + missing fact) and one OpenBookQA item as \textit{Question Ambiguity} alone. This demonstrates that the attribution module can reliably pinpoint the nature of uncertainty.
The remaining four cases (discussed in Section \ref{sec:5.2}) focus on Knowledge-Gap Extraction:
In two examples without heavy calculation (Figure \ref{fig:example2}, Figure \ref{fig:example1}), injecting the extracted conceptual fact fully resolves uncertainty and yields the correct answer with low entropy.
In two more complex examples (Figure \ref{fig:example4}, Figure \ref{fig:example3}) involving multi‐step numerical or physical–chemical reasoning, the model still errs after context injection. Nonetheless, in each case the extracted knowledge gap is precisely the missing formula or principle, confirming our module’s ability to localize exactly which piece of domain knowledge the model failed to apply.



\end{document}